\documentclass{article} % For LaTeX2e
\usepackage{iclr2026_conference,times}

\usepackage{amsmath,amsfonts,bm}

\def\eqref#1{equation~\ref{#1}}
\def\1{\bm{1}}

\DeclareMathAlphabet{\mathsfit}{\encodingdefault}{\sfdefault}{m}{sl}
\SetMathAlphabet{\mathsfit}{bold}{\encodingdefault}{\sfdefault}{bx}{n}

\usepackage{mmstyle}

\usepackage{hyperref}
\usepackage{url}
\usepackage{graphicx}

\usepackage{CJK}
\usepackage{caption}
\usepackage{graphicx}
\usepackage{multicol}
\usepackage{multirow}
\usepackage{booktabs}
\usepackage{amsmath}
\usepackage{upgreek}
\usepackage{footmisc}
\usepackage{tablefootnote}
\usepackage{footnote}
\usepackage{tabularx}
\usepackage{diagbox}
\usepackage{fancyvrb}
\usepackage{spverbatim}
\usepackage{nicematrix}
\usepackage{colortbl}
\usepackage{pythonhighlight}
\usepackage{listings}
\usepackage{amsmath}
\usepackage{amssymb}
\usepackage{mathtools}
\usepackage{amsthm}
\usepackage{multirow}
\usepackage{lscape}
\usepackage{listings}
\usepackage{subfigure}
\usepackage[normalem]{ulem}
\useunder{\uline}{\ul}{}

\theoremstyle{plain}

\theoremstyle{definition}

\theoremstyle{remark}

\usepackage{stfloats}

\usepackage{wrapfig}


\usepackage{framed}
\usepackage{makecell}
\usepackage{listings}
\definecolor{lightgray}{rgb}{.9,.9,.9}
\definecolor{darkgray}{rgb}{.4,.4,.4}
\definecolor{purple}{rgb}{0.65, 0.12, 0.82}
\lstdefinelanguage{JavaScript}{
  keywords={break, case, catch, continue, debugger, default, delete, do, else, false, finally, for, function, if, in, instanceof, new, null, return, switch, this, throw, true, try, typeof, var, void, while, with},
  morecomment=[l]{//},
  morecomment=[s]{/*}{*/},
  morestring=[b]',
  morestring=[b]",
  ndkeywords={class, export, boolean, throw, implements, import, this},
  keywordstyle=\color{blue}\bfseries,
  ndkeywordstyle=\color{darkgray}\bfseries,
  identifierstyle=\color{black},
  commentstyle=\color{purple}\ttfamily,
  stringstyle=\color{red}\ttfamily,
  sensitive=true
}
\usepackage{xspace}
\usepackage[shortlabels]{enumitem}

\PassOptionsToPackage{unicode}{hyperref}
\usepackage[most]{tcolorbox}
\tcbset{
  aibox/.style={
    width=390.18663pt,
    top=10pt,
    colback=white,
    colframe=black,
    colbacktitle=black,
    enhanced,
    center,
    attach boxed title to top left={yshift=-0.1in,xshift=0.15in},
    boxed title style={boxrule=0pt,colframe=white,}
  }
}
\newtcolorbox{AIbox}[2][]{aibox,title=#2,#1}
\tcbuselibrary{breakable,skins,theorems}
\newtcolorbox{tldrnbox}{
  enhanced,
  breakable,
  colback=black!2,      % 背景：很淡的灰
  colframe=black!18,    % 边框：淡灰
  boxrule=0.6pt,
  arc=2mm,
  left=1.5mm,right=1.5mm,top=1mm,bottom=1mm,
}

\newcommand\methodname{TAIL}
\newcommand\moduleState{Atomic State}
\newcommand\moduleTape{Memory Fetcher}
\newcommand\moduleExpan{Linear Transition}

\title{The Imitation Game: Turing Machine Imitator is Length Generalizable Reasoner}

\author{Zhouqi Hua$^{1,2*}$\quad Wenwei Zhang$^{2*\dag}$\quad Chengqi Lyu$^{2}$\quad Yuzhe Gu$^{2,3}$\quad \\\vspace{4pt}\textbf{Songyang Gao$^{2}$\quad Kuikun Liu$^{2}$\quad Dahua Lin$^{2,4}$\quad Kai Chen$^{2\dag}$} \\
$^1$Fudan University\quad $^2$Shanghai AI Laboratory\quad $^3$Shanghai Jiao Tong University \quad \\
$^4$MMLab, The Chinese University of Hong Kong \\
\texttt{\{huazhouqi,zhangwenwei,chenkai\}@pjlab.org.cn}
}

\iclrfinalcopy % Uncomment for camera-ready version, but NOT for submission.
\begin{document}

\maketitle

% !TEX root = ../neurips_2024.tex
\begin{abstract}
Length generalization, the ability to solve problems of longer sequences than those observed during training, poses a core challenge of Transformer-based large language models (LLMs).
Although existing studies have predominantly focused on data-driven approaches for particular arithmetic operations or symbolic manipulation tasks, these approaches tend to be task-specific with limited performance on individual tasks.
To pursue a more general solution, this paper focuses on a broader classes of reasoning problems that are \textit{computable}, \ie, problems that algorithms can solve, thus can be solved by the Turing machine, which operates over inputs of unbounded length.
From this perspective, this paper proposes \textbf{T}uring m\textbf{A}chine \textbf{I}mitation \textbf{L}earning \textbf{(\methodname{})} to improve the length generalization ability of LLMs.
\methodname{} uses computer programs to directly synthesize chain-of-thought (CoT) data that imitate the execution process of a Turing machine, which \textit{linearly} expands the reasoning steps into \textit{atomic} states to alleviate shortcut pattern learning and explicit \textit{memory} fetch mechanism to reduce the difficulties of dynamic and long-range data access.
To validate the universality and reliability of \methodname{}, we construct a challenging synthetic dataset covering 8 classes of algorithms and 18 tasks.
With only synthetic data, \methodname{} significantly improves the length generalization ability as well as the performance of Qwen2.5-7B in individual tasks, surpassing previous data-driven methods and DeepSeek-R1.
The experimental results reveal that the key concepts in the Turing machine, instead of the human-like thinking styles, are indispensable for \methodname{} for length generalization, through which the model exhibits read-and-write behaviors consistent with the properties of the Turing machine in their attention layers. This work provides a promising direction for future research in the learning of LLM reasoning from synthetic data.
% Our results also reveal that the thinking styles and reinforcement learning (RL), which have recently been believed to be important for generalization in large reasoning models, are unnecessary.
\end{abstract}

% \begin{figure}[h]
%     \centering
%     \includegraphics[width=1.0\linewidth]{./figures/head_image.pdf}
%     \caption{The average pass@1 accuracy of InternThinker-V 38B and some open-source models across eight reasoning benchmarks, as well as their pass@1 accuracy on OlympiadBench, MathVision test and MMMU Pro. Results show that InternThinker-V 38B has significantly improved over its base model, InternVL 2.5 38B (Instruct), achieving state-of-the-art performance on these benchmarks.}
%     \vspace{-16pt}
%     \label{fig:main_perf}
% \end{figure}
% !TEX root = ../neurips_2024.tex

\section{Introduction}
\label{sec:intro}

% Para 1 : intro to LG and challenge

Length generalization~\citep{press2021train}, \ie, the ability to handle a problem with input sequences of various lengths in the open world, especially those \textit{longer} than previously seen, is a fundamental aspect of human intelligence and serves as a crucial evaluation criterion for AI systems~\citep{anil2022exploring, sinha2024survey, ahuja2024provable, shi2022compositional}. 
Although the ability and generalizability of large language models (LLMs) to solve complex problems have been significantly improved by chain-of-thought (CoT)~\citep{wei2022chain}, recent studies~\citep{saparov2022language,anil2022exploring,zhou2024transformers} indicate that LLMs still struggle with length generalization, which sometimes explores and falls into shortcuts that eventually cause errors~\citep{saparov2024transformers}.

% Para 2 : existing methods problems + question arise

To address the challenge, existing works~\citep{zhou2024transformers,zhou2023algorithms,lee2023teaching,shen2023positional,mcleish2024transformers} primarily focus on data-driven approaches, which refine the training data by 
modifying the structure of CoT to be more effective and generalizable.
However, these methods remain inherently task-specific, \eg, Index Hint~\citep{zhou2024transformers,zhou2023algorithms} for symbolic reasoning tasks and Reversed Format~\citep{lee2023teaching,shen2023positional,zhou2023algorithms,mcleish2024transformers} for arithmetic problems, and yield only moderate performance gains. Thus, a question arises: \textit{Is there a universal and effective CoT structure for length generalization?}

\begin{figure}[t]
    \centering
    \makebox[\textwidth][c]{
        \includegraphics[width=1\linewidth]{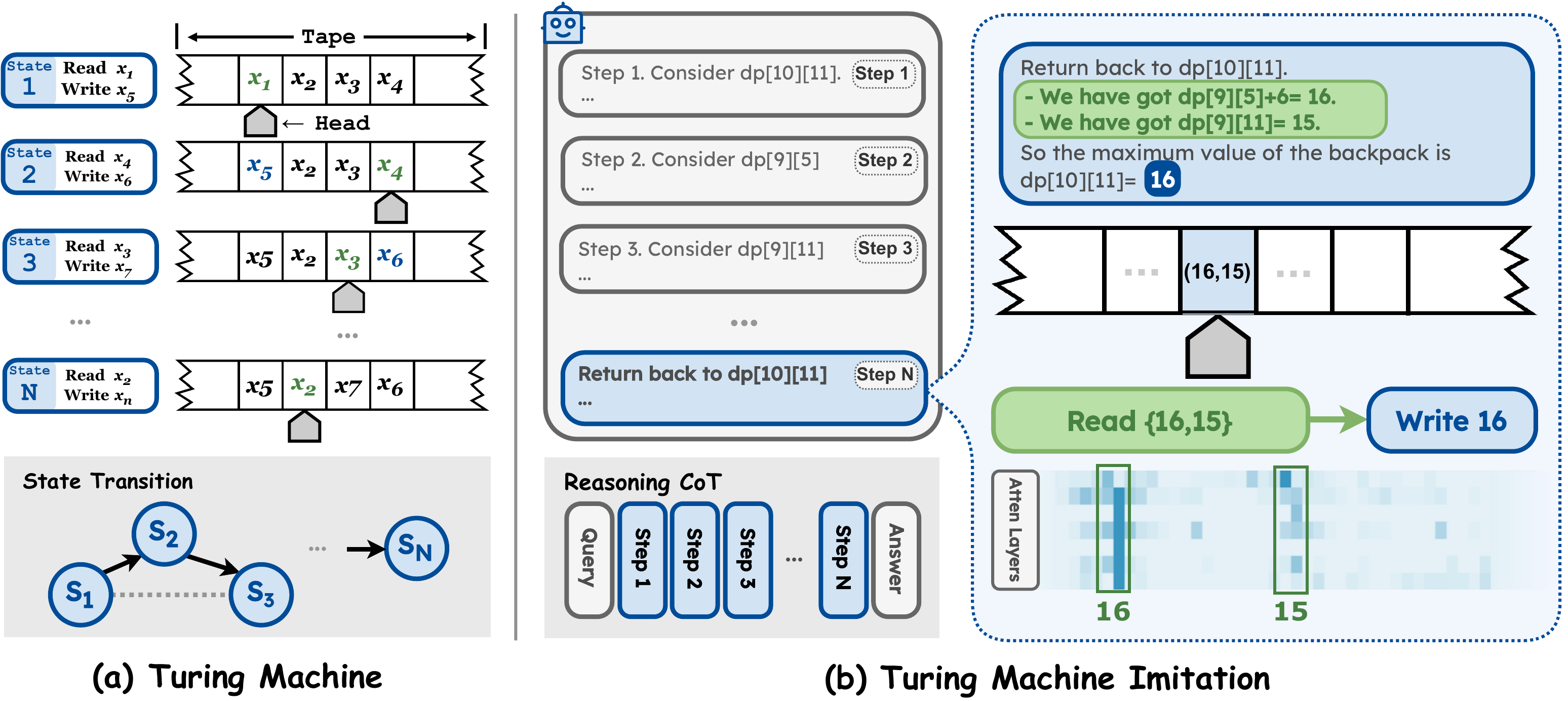}
    }
    \vspace{-5pt}
    \caption{Turing machine and its imitation in LLMs. \textbf{(a)} Illustration of a Turing machine performing algorithmic execution over a symbolic tape via sequential state transitions.
\textbf{(b)} TAIL simulates Turing machine execution by linearly structuring CoT into atomic read-write steps. Attention maps reveal operand retrieval and memory update patterns analogous to symbolic computation.}
    \vspace{-10pt}
    \label{fig:intro_attention_map}
\end{figure}

% Para 3 : Problem range definition + imitation goal -> algorithm
This paper aims to answer this question by first taking a deeper look at the commonalities among the problems. Notably, we observe that many of these tasks admit well-defined stepwise procedures that can be solved by program algorithms that generalize to inputs of arbitrary length. We refer to such tasks as \textbf{Computable Problems}, which serve as the focus of investigation in this paper. Thus, the core of achieving length generalization lies in letting LLMs faithfully simulate the execution process of the corresponding programs within their CoT for each problem.
In essence, the LLM acts like a Turing machine (Figure~\ref{fig:intro_attention_map}a), performing a sequence of fundamental operations on a memory tape, guided by finite states and logical transitions.

%Notably, we observe that many of these tasks admit well-defined stepwise procedures that can be finitely described and deterministically executed, suggesting that their solutions can be realized through discrete, symbolic transformations under bounded computational rules~\citep{turing1936computable,sipser1996introduction,arora2009computational,boolos2002computability}, which are referred as \textbf{Computable Problems}, solvable by algorithms that generalize to inputs of arbitrary length. 

% Para 4 : Algorithm -> TM -> Core module
%% Step(1) Algorithm -> TM
% According to the Church-Turing thesis~\citep{copeland1997church}, any algorithmically solvable problem can be computed by a , providing a new perspective for constructing a universal framework of .
% From this perspective
%% Step(2) Operation process of TM
% In each state, the head reads a unit of data from the infinite tape, writes a new value, and moves to the next position. Owing to the state transition table, even loops and other control flows can be unfolded into a linear sequence of states without any step skipping.
%% Step(3) TM -> Core module of TAIL
From this perspective, we propose \textbf{T}uring m\textbf{A}chine \textbf{I}mitation \textbf{L}earning (\textbf{\methodname{}}), which contains the three key structures in the synthesized CoT data that emulate three core properties of Turing machine execution: \moduleExpan{}, \moduleState{}, and \moduleTape{}.
First, similar to the Turing machine execution process, \textbf{\moduleExpan{}} enforces a complete and linear arrangement of reasoning steps to eliminate potential shortcut learning.
Second, \methodname{} decomposes the reasoning content into minimal units, termed \textbf{\moduleState{}s} to reduce difficulty and further reduce shortcut learning, which essentially correspond to the states of a Turing Machine, including read, write, and logical control operations.
Third, because LLMs can only append instead of modify in-place the tokens in their context due to their auto-regressive nature, the context of LLMs, which essentially serves as a memory, will keep growing as the reasoning continues. This poses difficulties for LLMs because of their attention mechanisms when they need to conduct elementary operations on operands that have long and dynamic distances among them. Therefore, \methodname{} further adopts a mechanism, termed \textbf{\moduleTape{}}, to read the necessary operand data and explicitly output them in the current step before conducting elementary operations.

% First, \methodname{} decomposes the overall reasoning trajectory into \textbf{\moduleState{}} to reduce the difficulty of each reasoning step, analogous to the single state in a Turing machine. Second, \textbf{\moduleTape{}} explicitly outputs the results in previous reasoning steps as operands for the current step, simulating data retrieval from the tape in a Turing machine, which reduces the difficulty of focusing on dynamic and long-range operand data in \moduleState{}. Third, \textbf{\moduleExpan{}} enforces a strictly sequential unfolding of the reasoning steps (Figure~\ref{fig:intro_attention_map}(b)), where each step corresponds to either an atomic operation, memory fetching, or a logic control, eliminating any possibility of step skipping or shortcut learning.
%% Step(4) cue 回图中
% , the linear transition of atomic reasoning operations containing explicit memory fetching constitutes a complete reasoning CoT, which ensures that it can be universal in a wide range of tasks at a higher level of abstraction than the task level.

% Para 5 : Experiments res
To assess the universality and effectiveness of \methodname{}, we construct a challenging dataset spanning 18 tasks across 8 algorithms, substantially harder than those in prior length generalization studies. Fine-tuning Qwen2.5-7B~\citep{yang2024qwen2} on this dataset yields high label accuracy across length ranges, with consistent gains on longer sequences, demonstrating strong length generalization over difficult samples. The model outperforms prior methods~\citep{zhou2024transformers,lee2023teaching,shen2023positional,zhou2023algorithms,martinez2023combining,mcleish2024transformers} and surpasses DeepSeek-R1~\citep{guo2025deepseek}.
Ablation studies show that removing any core module of \methodname{} severely degrades long-sequence performance. Notably, even minimalist CoT data containing only core modules without any thinking styles\footnote{\textit{Thinking styles} refers to the human-like linguistic expressions in CoT reasoning which are very common in existing large reasoning models, \ie, the surface-level natural language narrative rather than the underlying reasoning mechanism.} maintains full effectiveness, confirming \methodname{} as the key data-driven enabler of length generalization.
We also visualize the attention maps of the \methodname{}-fine-tuned model and observe that the attention during write operations focuses on fetched operands within the same state, resembling Turing machine behavior (Figure~\ref{fig:intro_attention_map}b).

% !TEX root = ../neurips_2024.tex

\section{Preliminaries}

% Part 1 : intro on LG

\subsection{Length Generalization and Computable Problems}

For large language models (LLMs), length generalization means that a model can process long input sequences, although it is only trained on short sequences. 
% It can be formalized as
% \begin{align}
% \label{eq:length_generalizaion}
% P_\theta(x_{\text{test}} | L_{\text{test}} > L_{\text{train}}) = f_\theta(x_{\text{train}}).
% \end{align}
% where $L_{\text{train}}$ and $L_{\text{test}}$ represent the input sequence lengths 
% and $f_\theta$ means that the model still uses the parameters $\theta$ trained on short sequences. 
For example, a model trained on 10-30 digit addition can maintain strong performance on 30-50 digit addition tasks.
Fundamentally, successful length generalization implies that the model has extracted a structural pattern from the training data. This pattern should be general and can scale adaptively with input length.
% When the reasoning problem becomes more complex, this process can be unfolded into multi-step reasoning with the help of Chain-of-Thought (CoT), 
% which can be represented as
% \begin{align}
% \label{eq:CoT_reasoning}
% P_\theta(x_{\text{test}}) = \prod_{i=1}^{n} P(x_i | x_{<i}, \theta),
% \end{align}
% where $n$ represents the total steps of reasoning. Each single reasoning step can be formalized as $P(x_i | x_{<i}, \theta)$, using previous reasoning results $x_{<i}$ to predict the current result $x_i$. Thus, the complete reasoning path can be expressed as $(x_0, x_1, \dots, x_n)$. 
% After formalizing the problem of length generalization, we shift our focus to the reasoning tasks themselves, conducting a systematic analysis. 

After a deeper look at the problem, incorporating insights from prior indirect conclusions~\citep{deletang2022neural}, we observe that many tasks can essentially be solved through discrete symbolic transformations governed by bounded algorithmic computational rules~\citep{turing1936computable,sipser1996introduction,arora2009computational,boolos2002computability}. For example, Parity can be solved through a simple enumeration procedure, while arithmetic addition can be handled by simulating the full digit-wise addition process, including carry propagation. We refer to such tasks as \textit{Computable Problems}, whose commonality lies in being solvable by a well-defined, deterministic algorithmic procedure. 
Such algorithms inherently handle inputs of arbitrary length, which aligns with the goal of length generalization. Training LLMs to learn their step-by-step execution thus enables generalization across input lengths when solving computable problems.

\subsection{Turing Machine}
\label{subpara:TM_Definition}
While all computable problems are solvable by algorithms, their structural diversity makes chain-of-thought (CoT) design impractical.
Therefore, a more abstract and general framework is essential to unify the CoT paradigm for computable problems.
Based on the Church-Turing thesis~\citep{copeland1997church}, a Turing machine can solve any algorithmically computable problem, thereby providing a universal and higher-level framework for problem solving.
In other words, the computational trace data of any computable problem can be constructed by simulating the execution of a Turing machine.

The formal definition of the Turing machine~\citep{turing1936computable,hopcroft2001introduction} consists of an infinite-length tape, a read/write head, and a table containing a finite set of state transitions. It can be represented as a 7-tuple:
\begin{align}
\label{eq:Pre_7tuple}
    M = (Q, \Sigma, \Gamma, \delta, q_0,B,F),
\end{align}
where $Q$ is a finite set of states, $\delta$ is the transition function, $q_0 \in Q$ refers to the initial state (see Appendix~\ref{sec:def_TM} for full definitions). In any non-accepting state $q_s\in Q$, the head reads a symbol $a$ from the tape, overwrites it with a new symbol $b$, and moves the head to a new position, thereby transitioning to the next state $q_{s+1}$, which can be formally defined as:
\begin{align}
\label{eq:Pre_main}
\delta(q_s, a) = (q_{s+1}, b, D),
\end{align}
where the head moves one position in direction $D$. Thus, $\delta$ represents a complete state transition conflating two logically independent states, and a linear unfolding of states $q_0 \to q_1 \to \dots \to q_n$ represents the complete process of Turing machine implementing the program. In order to align the reasoning process of LLMs with Turing Machine, the reasoning procedure can be unfolded into multi-step reasoning with the help of CoT.
% which can be represented as:
% \begin{align}
% \label{eq:CoT_reasoning}
% P_\theta(x) = \prod_{i=0}^{n} P(x_i | x_{<i}, \theta),
% \end{align}
% where $n$ represents the total steps of reasoning.
Each single reasoning step can be formalized as $x_i$ in CoT, deriving the current reasoning result based on the preceding reasoning steps $x_{<i}$. 
It is important to note that the granularity of $x_i$ is determined by the size of the reasoning step in the specific task. Typically, it corresponds to the prediction of multiple tokens, forming an intermediate reasoning outcome at each stage.
$x_0$ represents the query and thus the complete reasoning path (CoT) can be expressed as $x_0\to x_1\to \dots\to x_n$. In line with the Turing machine, each reasoning step $x$ corresponds to a Turing Machine state $q$ in Eq.(\ref{eq:Pre_main}), which includes reading an input symbol $a$. The entire CoT is formed by a linear composition of such steps, analogous to the full unrolling of the Turing Machine’s state transitions $\delta$ from $q_1$ to $q_n$. 

Previous work~\citep{li2024chain} has theoretically shown that Transformers can achieve Turing Completeness given sufficiently long CoT, but has not provided concrete guidelines for constructing such CoT sequences in a wide range of tasks.
Based on our analysis, multi-step reasoning CoT can be structurally aligned with the computation process of a Turing machine. This leads us to hypothesize that, by endowing the reasoning process of LLMs with key properties of a Turing machine, the model can effectively simulate algorithmic execution and achieve length generalization.

% Recent work~\citep{li2024chain} demonstrates that Transformers are Turing-complete given sufficiently long chains of thought, which makes it feasible for LLMs to simulate a Turing machine. Therefore, Eq.(\ref{eq:Pre_main}) serves as the reference for our TAIL framework.

% !TEX root = ../neurips_2024.tex

% \newcommand\moduleState{Atomic Operation}
% \newcommand\moduleTape{Memory Fetcher}
% \newcommand\moduleExpan{Linear Transition}

% arc overview & our CoT & other's poor CoT
\begin{figure}[t]
    \makebox[\textwidth][c]{
        \includegraphics[width=1\linewidth]{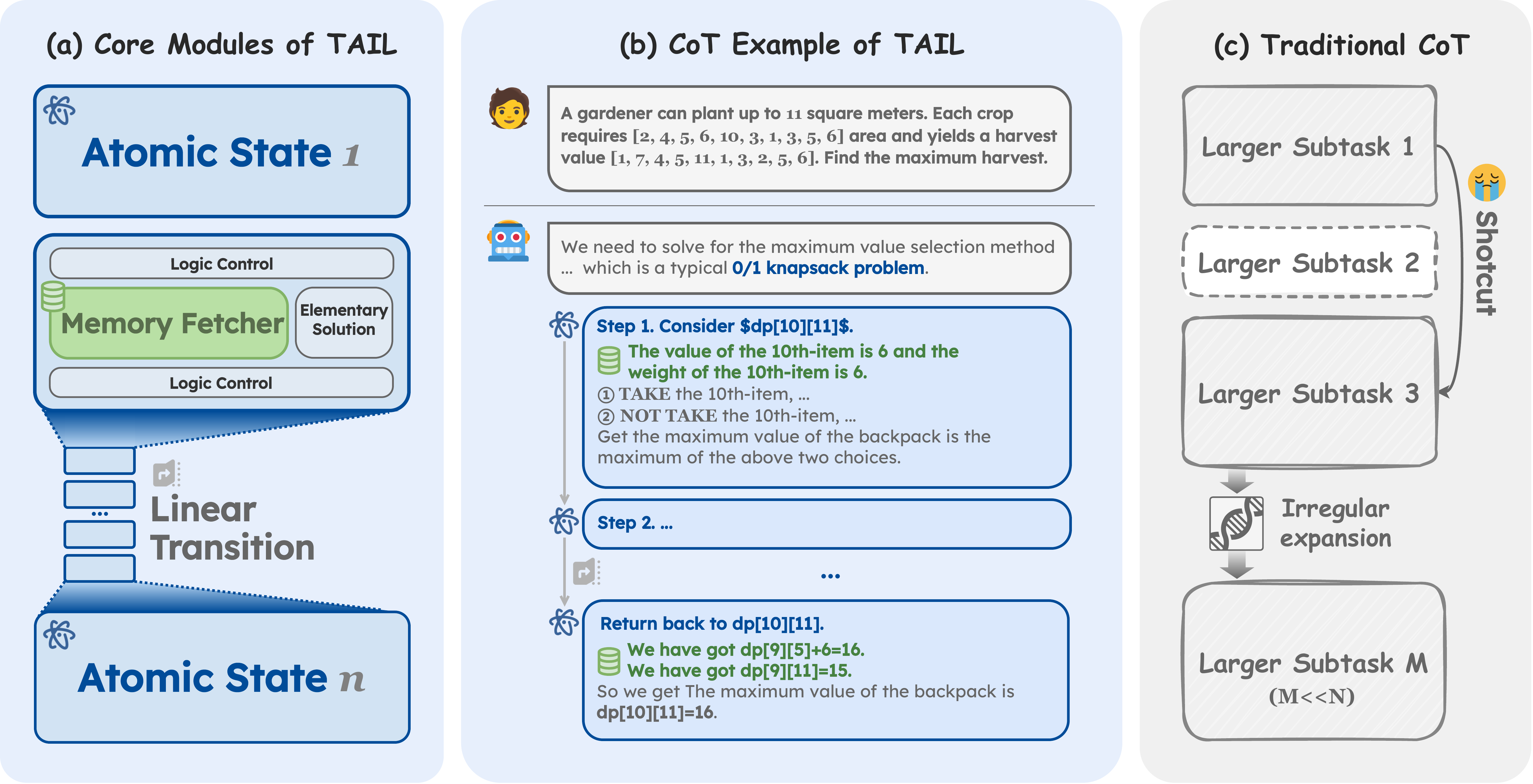}
    }
    \vspace{-10pt}
    \caption{An overview of \methodname{}. \textbf{(a) Core Modules of \methodname{}} imitate a Turing Machine, containing a \moduleExpan{} of \moduleState{} with \moduleTape{} of previous reasoning results. \textbf{(b) CoT generated by \methodname{}}: the solution to a 0/1 knapsack problem using a dynamic programming algorithm. \textbf{(c) Traditional CoT} consists of oversized subtasks, shortcut learning, and irregular expansion.}
    \vspace{-10pt}
    \label{fig:overview}
\end{figure}

\section{Turing Machine Imitation Learning}
\label{sec: methods}

Based on the preceding analysis, this paper proposes \textbf{Turing mAchine Imitation Learning (\methodname{})} to align the Chain-of-thought (CoT) of large language models (LLMs) to simulate the execution of a Turing machine for achieving universal and effective length generalization. \methodname{} imitates key properties of a Turing Machine (Figure~\ref{fig:overview}), comprising three core modules spanning macro to micro levels: {\moduleExpan}
(Section~\ref{subpara:expansion}), {\moduleState}
(Section~\ref{subpara:atomic}) and {\moduleTape}
(Section~\ref{subpara:review}).

\subsection{\moduleExpan}
\label{subpara:expansion}

According to the RASP-Generalization Conjecture~\citep{zhou2023algorithms}, Transformer-based LLMs struggle with problems that involve intricate control structures, such as loops. This suggests the need to transform these structures into simpler forms that align better with the capabilities of the model. In particular, complex reasoning structures (like trees and graphs) can be linearly unrolled and traversed to enable complete and non-redundant execution of all reasoning steps, thereby preventing shortcuts in the reasoning process.
Similarly, in a Turing machine, the execution of a complete program corresponds to a linear unfolding of states $q_1 \to q_2 \to \dots \to q_n$ as shown in Eq.(\ref{eq:Pre_main}), where even control structures such as loops can be flattened into a sequential process.
To align with this characteristic, we introduce {\moduleExpan}, which describes from a macro-level perspective how individual reasoning steps are composed into a linear and orderly structure within the overall reasoning process, and collectively form the CoT.

% {\moduleExpan} describes how individual reasoning steps are composed within the overall reasoning process from a macro-level perspective. Analogous to a Turing machine, where the execution of a complete program corresponds to a linear unfolding of states $q_1 \to q_2 \to \dots \to q_n$ in E.q.(\ref{eq:Pre_main}), even control structures such as loops can be flattened into a sequential process. This ensures strictly step-by-step execution and prevents any form of state skipping.
% We imitate this property within CoT by ensuring that all individual reasoning steps are sequentially connected through tightly constrained logical dependencies, \ie, $x_1\to x_2\to \dots \to x_n$ in E.q.(\ref{eq:CoT_reasoning}).

\subsection{\moduleState}
\label{subpara:atomic}

Although \moduleExpan{} defines the overall structure of CoT reasoning as a linear sequence of reasoning steps, it does not impose constraints on the size of each step. Overly large reasoning steps not only increase the difficulty of learning for the model but also risk introducing shortcuts within a single step.
Therefore, we attempt to constrain the size of a reasoning step by enforcing a standardized internal structure.
Inspired by the Turing machine, each state encompasses a sequence of simple operations: \textit{reading} data from the tape, \textit{writing} new data, and \textit{transitioning} to the next state. Following this principle, we define {\moduleState} consisting of operand retrieval (realized via {\moduleTape}, detailed in Section~\ref{subpara:review}), the elementary solution produced within the reasoning step, and a set of logical control statements, as shown in Figure~\ref{fig:overview}(a).
Meanwhile, following the RASP-L hypothesis\footnote{We do not strictly follow RASP-L to constrain each reasoning step, but use it to indicate problems directly solvable by Transformers. This relaxed view shows strong length generalization in our experiments.}, we argue that each \moduleState{} should adhere to the principles of realizability and simplicity. Specifically, since we use Python programs to synthesize CoT, we define an \moduleState{} as a single algorithmic step in the program without internal loops.

\subsection{\moduleTape}
\label{subpara:review}

\begin{figure}[t]
    \centering
    \includegraphics[width=1\linewidth]{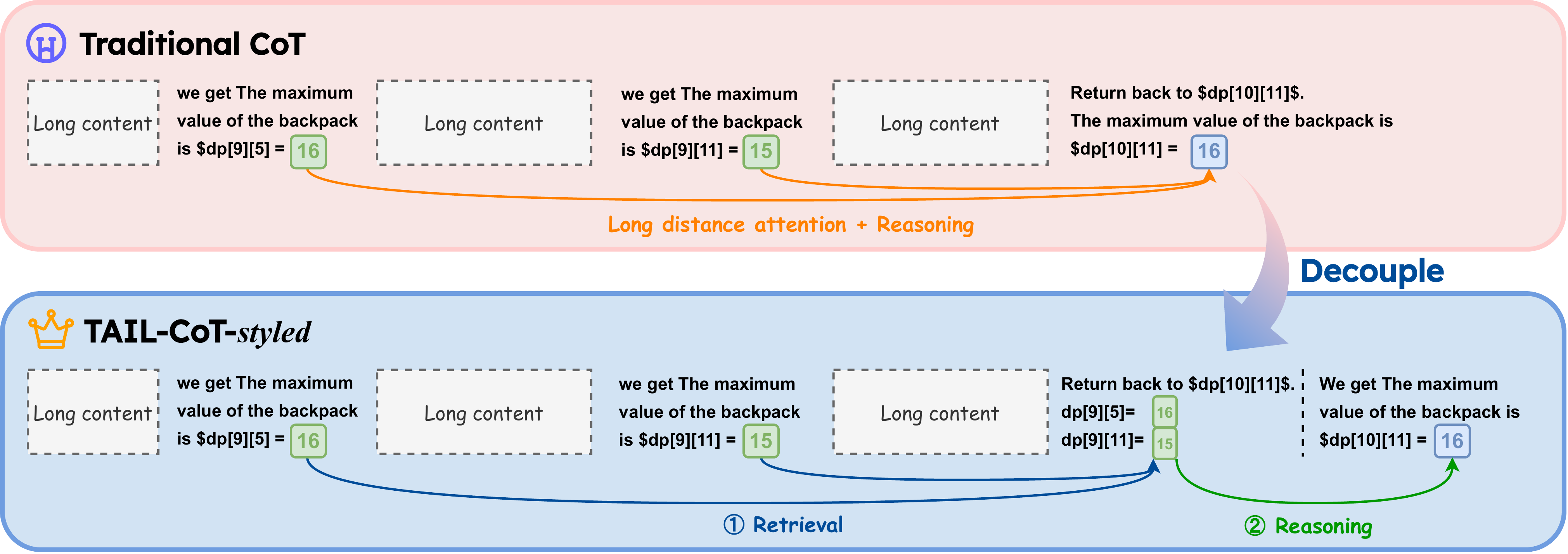}
    \caption{Comparison of traditional CoT and \methodname{}'s \textit{Memory Fetcher}, which decouples long-distance attention and the reasoning process.}
    \label{fig:MF_cmp}
    \vspace{-10pt}
\end{figure}

In a Turing machine, every state reads data from the tape and replaces it with processed result. However, auto-regressive models (\eg, Transformer) can only extend the token sequence by appending new tokens instead of in-place token modification.
% , such as simulating result writing by appending the elementary solution, and guiding state transitions through explicit logic control. 
So the action of data reading is typically achieved by constructing attention mechanisms over previous tokens. As reasoning progresses, the sequence grows longer, requiring the model to retrieval over increasingly distant and dynamically shifting tokens. Furthermore, simultaneously performing data retrieval and generating the elementary solution at the same time increases the learning difficulty for the model. 
% To address this, we propose \moduleTape{}, which decouples the act of establishing long-range attention from the actual execution of reasoning operations by:
To address this, we propose \moduleTape{} to decouple these two operations by:
(1) \textit{first} explicitly outputting all relevant operands at the beginning of every \moduleState{}, (2) \textit{then} performing reasoning and outputting local results.
As shown in an example in Figure~\ref{fig:MF_cmp}, \moduleTape{} changes the attention structure by localizing operands and improves reasoning accuracy, which has been theoretically proved by recent work~\citep{wang2025beyond}.
Figure~\ref{fig:res_attention} compares the attention structures with and without \moduleTape{}. It is obvious that \moduleTape{} enables precise localization of relevant operands during reasoning through prominent local attention. See more details in Appendix~\ref{sec:atten_vis}.
\section{Experiment}
\label{sec:experiment}

\subsection{Dataset Synthesis}

\noindent\textbf{Task Selection.}
This work focuses on length generalization in hard samples rather than unlimited extension in simple tasks.
So we synthesize a set of \textit{challenging} tasks based on 8 classic algorithmic paradigms in computable problems
% (\ie, Simulation, Recursion, Iteration, Greedy, Enumeration, DP, Divide \& Conquer and Backtracking) 
to verify the effectiveness of \methodname{}. As shown in Table~\ref{tab:algorithm_task_length}, the dataset comprises 18 tasks, including previously studied problems such as addition, but with randomized digit lengths and decimal places to increase difficulty. Each task has a high degree of diversity in query narratives, some of which incorporate real-world problems (\eg, Diophantine Equation, 0-1 Knapsack, etc).

\begin{wraptable}{lt}{0.48\columnwidth}
% \vspace{-6pt}
\centering
\fontsize{8}{7}\selectfont
\renewcommand{\arraystretch}{1.2}
\setlength{\tabcolsep}{3pt}

\begin{tabularx}{\linewidth}{@{} l *{3}{>{\centering\arraybackslash}X} @{}}
\toprule
\textbf{Method} & \textbf{S} & \textbf{M} & \textbf{L} \\
\midrule
Index Hint & 57.0 & 34.5 & 24.0 \\
Reversed Format & 39.5 & 35.5 & 35.0 \\
\rowcolor[HTML]{E6F2FF}
\textbf{\methodname{} (Ours)} & \textbf{97.0} & \textbf{92.5} & \textbf{86.5} \\
\bottomrule
\end{tabularx}

% \vspace{2pt}
\caption{Pass@1 (\%) on large-number addition, compared to previous data-driven approaches.}
\label{tab:model_comparison}
\vspace{-8pt}
\end{wraptable}

\noindent\textbf{Synthesis Approach.}
We employed supervised fine-tuning (SFT) with synthetic data to internalize the model's ability to generate Chain-of-Thought (CoT) with \methodname{}'s core modules.
Figure~\ref{fig:data_flow} illustrates the data synthesis process of \methodname{}. We claim that \methodname{} is task \textit{universal}\footnote{Previous methods like Index Hint or Reversed Format, due to their structural specificity, cannot be effectively constructed for tasks beyond simple bit-matching operations.} because for each task belonging to a specific algorithm, it's feasible to construct a Python program and add string append statements to assemble CoT. When the program runs, the resulting CoT reflects the complete program execution flow.
We implement the injection of three core modules in CoT through the following methods: (1) Treating each algorithmic step as an \moduleState{}, especially each time entering a loop. (2) Unfolding the algorithmic process sequentially as \moduleExpan{}, achieved by using programs to synthesize CoT itself. (3) Explicitly outputting all relevant operands of current algorithm step as \moduleTape{} in CoT.
During the generation process, we performed strict de-duplication and ensured that none of the data in the evaluation set was included in the training set. 
For training data, we first validated the sufficiency of the \methodname{} architecture by synthesizing \textbf{\methodname{}-CoT} that only includes three core modules in a format similar to that shown in Figure~\ref{fig:mini_cot}.
Then we enrich \methodname{}-CoT into \textbf{\methodname{}-CoT-\textit{styled}} using natural language (see Figure~\ref{fig:user_cot}) and verify length generalization ability on all 18 tasks, as shown in Figure~\ref{fig:res_dsr1}.
See Appendix~\ref{sec:Data_Flow} for more details.

\noindent\textbf{Dataset Size.}
To better facilitate training and evaluation of length generalization, we defined three length ranges for each task: Short (\textbf{S}), Medium (\textbf{M}) and Long (\textbf{L}). For comprehensive training, we synthesized 100,000 training samples and 500 evaluation samples for each length range, resulting in 1,500 evaluation samples per task. The validation of length generalization refers to whether a model trained on the S-range training set can avoid sharp performance degradation on the M- and L-range evaluation sets. The length ranges of each task are detailed in Table \ref{tab:task_length_range}.

\subsection{Experimental Settings}
\label{subsec:exp_setting}

\begin{figure}[t]
    \centering
    \includegraphics[width=1\linewidth]{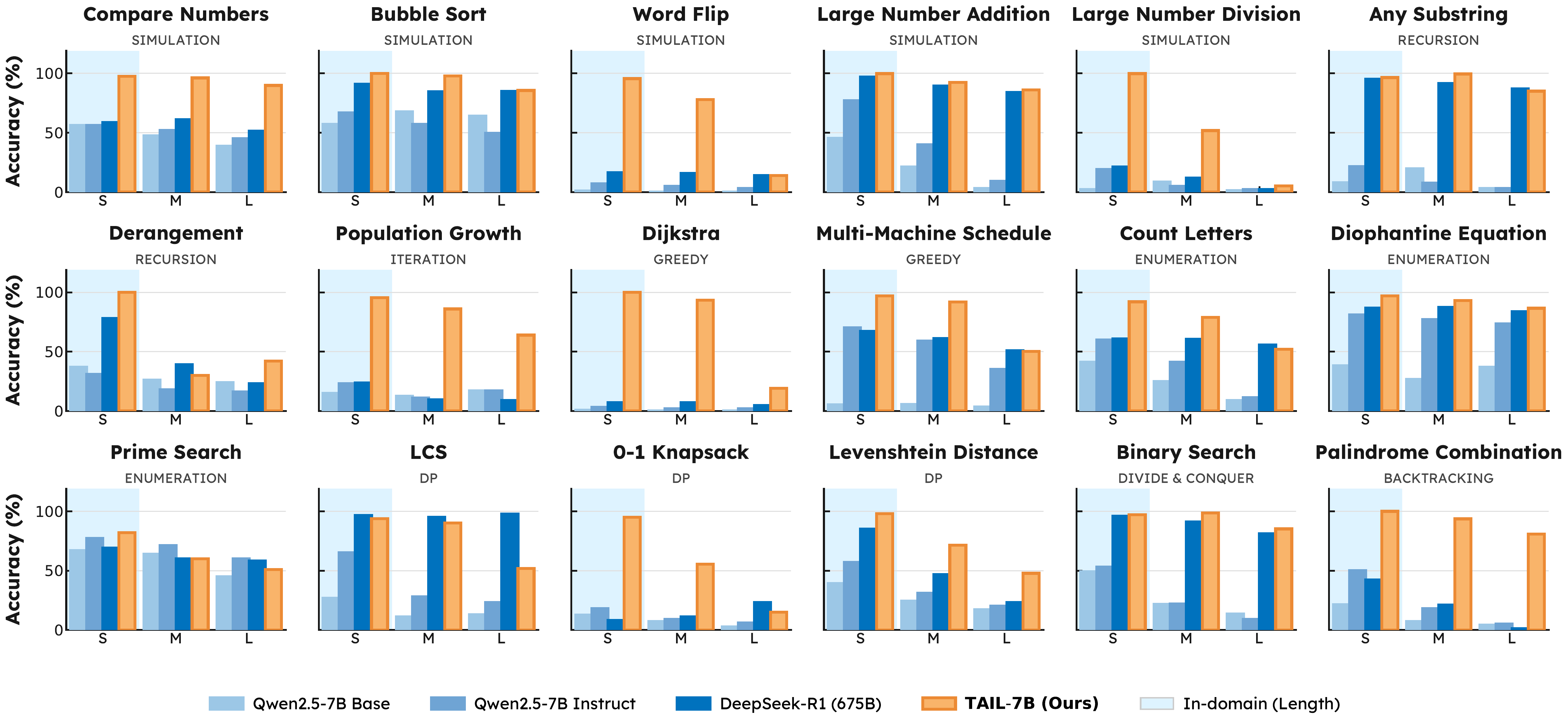}
    \caption{Length generalization performance of Qwen2.5-7B finetuned with \methodname{}-CoT-\textit{style} across all 18 tasks, in comparison with Qwen2.5-7B (base model), Qwen2.5-7B Instruct and DeepSeek-R1.} 
    \vspace{-10pt}
    \label{fig:res_dsr1}
\end{figure}

\noindent\textbf{Metrics.}
Previous work~\citep{saparov2022language} has demonstrated experimentally that \textit{label accuracy} is well suited to measure reasoning capability of LLMs. We use pass@1 label accuracy under the zero-shot setting and use greedy decoding to evaluate.
% Since we only compare final results, for problems requiring specific methods (such as Binary Search), we strictly limit the method in the prompt and indicate that it is for educational purposes\footnote{For the DeepSeek-R1 inference results, we sampled 30 items from each length range in each task for manual inspection, the results showed that these restrictions did indeed work.}.

\noindent\textbf{Training.}
We fine-tuned Qwen2.5-7B with training 2 epochs for most tasks and more epochs for a few more challenging ones with a global batch size of 1024. 
The initial learning rate was 1e-5, decaying to 7e-7, with a weight decay of 0.1.

\noindent\textbf{Evaluation.}
To facilitate a more efficient evaluation procedure, we follow a dual-model framework. First, a small 1.5B specialized model extracts the answers. Then, Qwen2.5-72B-Instruct performs evaluations, outputting \verb|\boxed{YES}| or \verb|\boxed{NO}| to represent the evaluation result. 
% Since our inference tasks have a single ground-truth answer, the evaluation model enforces strict content matching while allowing some flexibility in answer formatting.

\begin{figure}[t]
    \centering
    \vspace{-5pt}
    \includegraphics[width=0.9\linewidth]{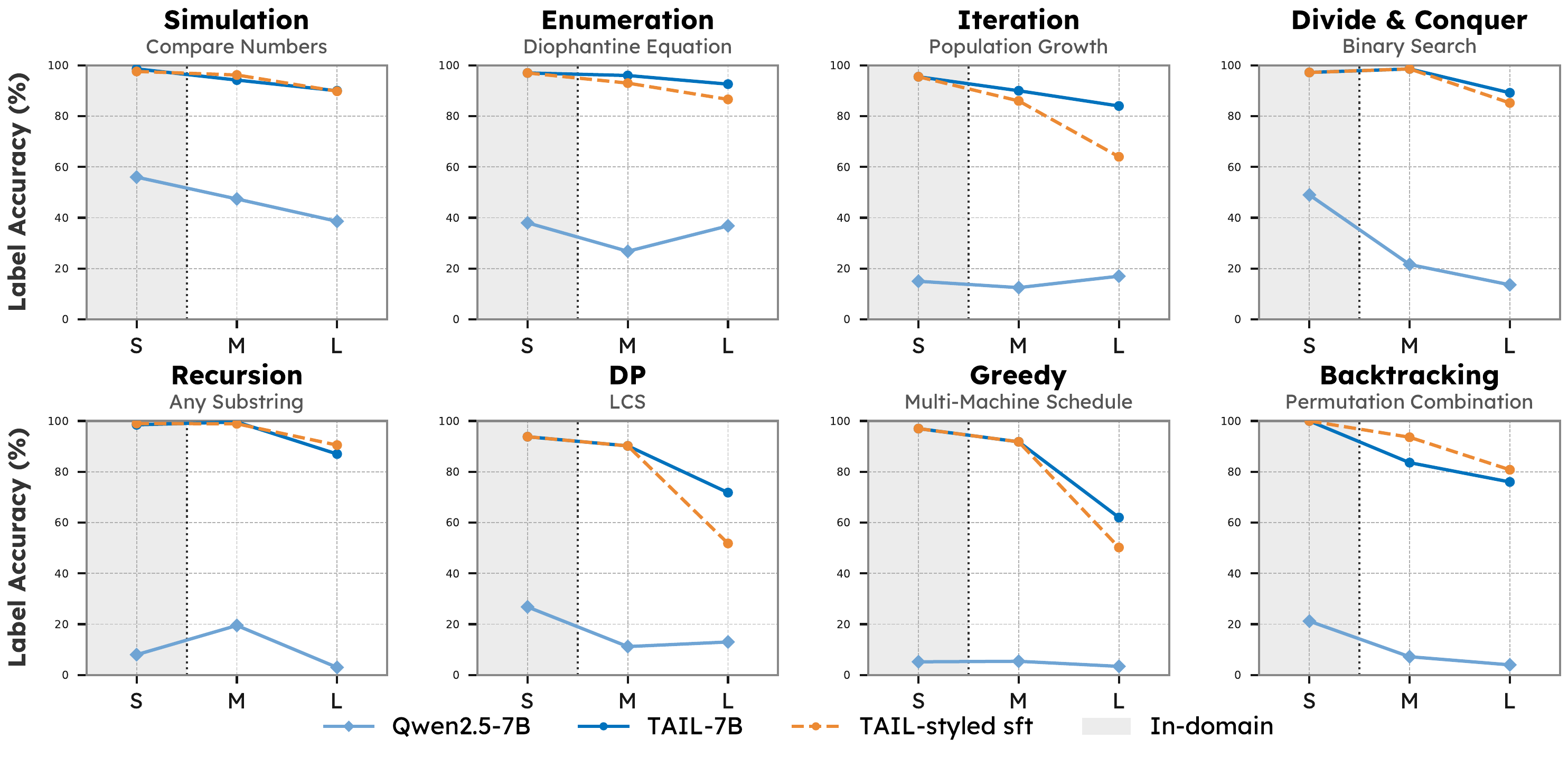}
    \caption{Comparison of fine-tuned Qwen2.5-7B with TAIL core module and the base model. For each algorithm, we select a representative task. After fine-tuning, model demonstrates length generalization on sequences that are 5 to 10 times longer than those in training.} 
    \vspace{-10pt}
    \label{fig:method_res}
\end{figure}

\begin{table}[t]
\centering
\fontsize{8}{7}\selectfont % 保持字体大小固定
\renewcommand{\arraystretch}{1.15} % 微调行距，不拥挤
\setlength{\tabcolsep}{1pt} % 基础间距，由 tabularx 自动拉伸

% 定义一个居中对齐的 X 列，用于数据列自动平分宽度
\newcolumntype{Y}{>{\centering\arraybackslash}X}

% ==================== 上半部分 (Part 1: 前4个算法) ====================
\begin{tabularx}{0.85\linewidth}{l *{8}{Y}}
\toprule
& \multicolumn{2}{c}{\textbf{Simulation}} 
& \multicolumn{2}{c}{\textbf{Enumeration}} 
& \multicolumn{2}{c}{\textbf{Iteration}} 
& \multicolumn{2}{c}{\textbf{Divide \& Conquer}} \\
\cmidrule(lr){2-3} \cmidrule(lr){4-5} \cmidrule(lr){6-7} \cmidrule(l){8-9}
\textbf{Model} & \textbf{M} & \textbf{L} & \textbf{M} & \textbf{L} & \textbf{M} & \textbf{L} & \textbf{M} & \textbf{L} \\
\midrule
Qwen2.5-7B Base & 47.4 & 38.6 & 26.8 & 36.8 & 12.4 & 17.0 & 21.6 & 13.6 \\
w/o Atomic State & 82.6 & 73.6 & 68.2 & 69.0 & 77.2 & 61.2 & 86.2 & 71.0 \\
w/o Linear Transition & 80.0 & 75.4 & 63.6 & 58.8 & 76.2 & 54.0 & 85.0 & 75.6 \\
w/o Memory Fetcher & 90.2 & 88.0 & 64.2 & 63.6 & 73.8 & 67.6 & 92.4 & 88.2 \\
\midrule
\rowcolor[HTML]{E6F2FF}
\methodname{} & \textbf{94.2} & \textbf{90.0} & \textbf{96.0} & \textbf{92.6} & \textbf{90.0} & \textbf{84.4} & \textbf{98.6} & \textbf{89.2} \\
\bottomrule
\end{tabularx}

\vspace{6pt} % 上下表格间距

% ==================== 下半部分 (Part 2: 后4个算法) ====================
\begin{tabularx}{0.85\linewidth}{l *{8}{Y}}
\toprule
& \multicolumn{2}{c}{\textbf{Recursion}} 
& \multicolumn{2}{c}{\textbf{DP}} 
& \multicolumn{2}{c}{\textbf{Greedy}} 
& \multicolumn{2}{c}{\textbf{Backtracking}} \\
\cmidrule(lr){2-3} \cmidrule(lr){4-5} \cmidrule(lr){6-7} \cmidrule(l){8-9}
\textbf{Model} & \textbf{M} & \textbf{L} & \textbf{M} & \textbf{L} & \textbf{M} & \textbf{L} & \textbf{M} & \textbf{L} \\
\midrule
Qwen2.5-7B Base & 19.4 & 3.0 & 11.2 & 13.0 & 5.4 & 3.4 & 7.2 & 4.0 \\
w/o Atomic State & 52.2 & 32.0 & 77.4 & 61.0 & 39.0 & 16.0 & 75.8 & 61.4 \\
w/o Linear Transition & 43.0 & 30.8 & 77.4 & 74.2 & 20.6 & 11.2 & 79.0 & 62.8 \\
w/o Memory Fetcher & 87.2 & 84.8 & 80.8 & 74.8 & 45.6 & 30.8 & 80.0 & 69.2 \\
\midrule
\rowcolor[HTML]{E6F2FF}
\methodname{} & \textbf{99.6} & \textbf{87.0} & \textbf{90.2} & \textbf{71.8} & \textbf{91.8} & \textbf{62} & \textbf{83.6} & \textbf{76} \\
\bottomrule
\end{tabularx}

\vspace{3px}
\caption{Ablation study in core modules of \methodname{}. For each algorithm, we select a representative task and evaluate pass@1 accuracy (\%) only on sequences that exceed the training length. Results show that the absence of any core module leads to a sharp degradation in length generalization performance.}
\label{tab:module_ablation}
\vspace{-15px}
\end{table}

\subsection{Performance}

\noindent\textbf{Overall Performance.}
% Due to its greater readability, we synthesized \methodname{}-CoT-\textit{styled} (as shown in Figure~\ref{fig:user_cot}) on all 18 tasks across 8 classes of algorithms, and fine-tuned Qwen2.5-7B. As shown in Figure~\ref{fig:res_dsr1}, We observe length generalization on most difficult tasks, where there was no sharp performance degradation on out-of-domain length sequences. Several tasks like \textit{Compare Numbers}, \textit{Bubble Sort} and \textit{Any Substring} reach near saturation in out-of-domain sequences. 
% Moreover, \methodname{} also outperformed Qwen2.5-7B (representing the base model), Qwen2.5-7B Instruct (representing fine-tuning on a large amount of traditional non-\methodname{}-CoT data), and DeepSeek-R1 671B (a representative open-source reasoning model) in both label accuracy and length generalization abilities. Compared with reasoning models (\ie, DeepSeek-R1), we conclude that the huge leap in performance lies in their different underlying mechanisms. Reasoning models often try many approaches but only scratch the surface and exploit shortcuts to bypass the structured reasoning process, instead of delving into a step-by-step approach (see more details in section~\ref{exp_reasoning_cmp}).
% However, we also find some limitations of \methodname{}, details can be seen in Appendix~\ref{dis}.
Due to its greater readability, we synthesized \methodname{}-CoT-\textit{styled} (as shown in Figure~\ref{fig:user_cot}) on all 18 tasks across 8 classes of algorithms, and fine-tuned Qwen2.5-7B. As shown in Figure~\ref{fig:res_dsr1}, We observe \textbf{length generalization on most difficult tasks}, where there was no sharp performance degradation on out-of-domain length sequences. Several tasks like \textit{Compare Numbers}, \textit{Bubble Sort} and \textit{Any Substring} reach near saturation in out-of-domain sequences. 
Moreover, \methodname{} also outperformed Qwen2.5-7B (base model), Qwen2.5-7B Instruct (fine-tuning on a large amount of traditional non-\methodname{}-CoT data), and DeepSeek-R1 671B (representative open-source reasoning model) in both label accuracy and length generalization abilities. Compared with reasoning models (\ie, DeepSeek-R1), we conclude that the huge leap in performance lies in their \textbf{different underlying mechanisms}. Reasoning models often try many approaches but only scratch the surface and exploit shortcuts to bypass the structured reasoning process, instead of delving into a step-by-step approach (see more details in section~\ref{exp_reasoning_cmp}).
Regarding concerns about CoT length, experiments show that TAIL-CoT achieves a significant performance leap with \textbf{slightly} \textbf{less token usage} compared to inference models, as shown in Appendix~\ref{sec:length_cmp}.

% Although \methodname{}-CoT(-styled) is longer than traditional CoT, the principle behind it is significantly different from that of reasoning models (\ie, DeepSeek-R1). 

\noindent\textbf{Comparison with Previous Works.}
Since previous methods (Index Hint~\citep{zhou2024transformers,lee2023teaching,shen2023positional,zhou2023algorithms,martinez2023combining,mcleish2024transformers} and Reversed Format~\citep{zhou2023algorithms,zhou2024transformers}) have proven effective on limited problems such as large number operations, we choose \textit{Large Number Addition} of \textit{Simulation} algorithm as a common task for comparison.
Unlike prior work using fixed-length integers, our setup samples two operands with random lengths and optional \textbf{decimal points}, greatly expanding the state space. 
We followed the method in previous works to construct the same amount of training data (see details in Appendix~\ref{sec:baseline_data}), and fine-tuned Qwen2.5-7B separately.
As shown in Table~\ref{tab:model_comparison}, models trained with Index Hint and Reversed Format under-perform \methodname{} by a large margin, highlighting the inadequacy of prior methods in addressing the challenges of length generalization in difficult tasks.

\begin{figure}[t]
    \centering
    \includegraphics[width=1\linewidth]{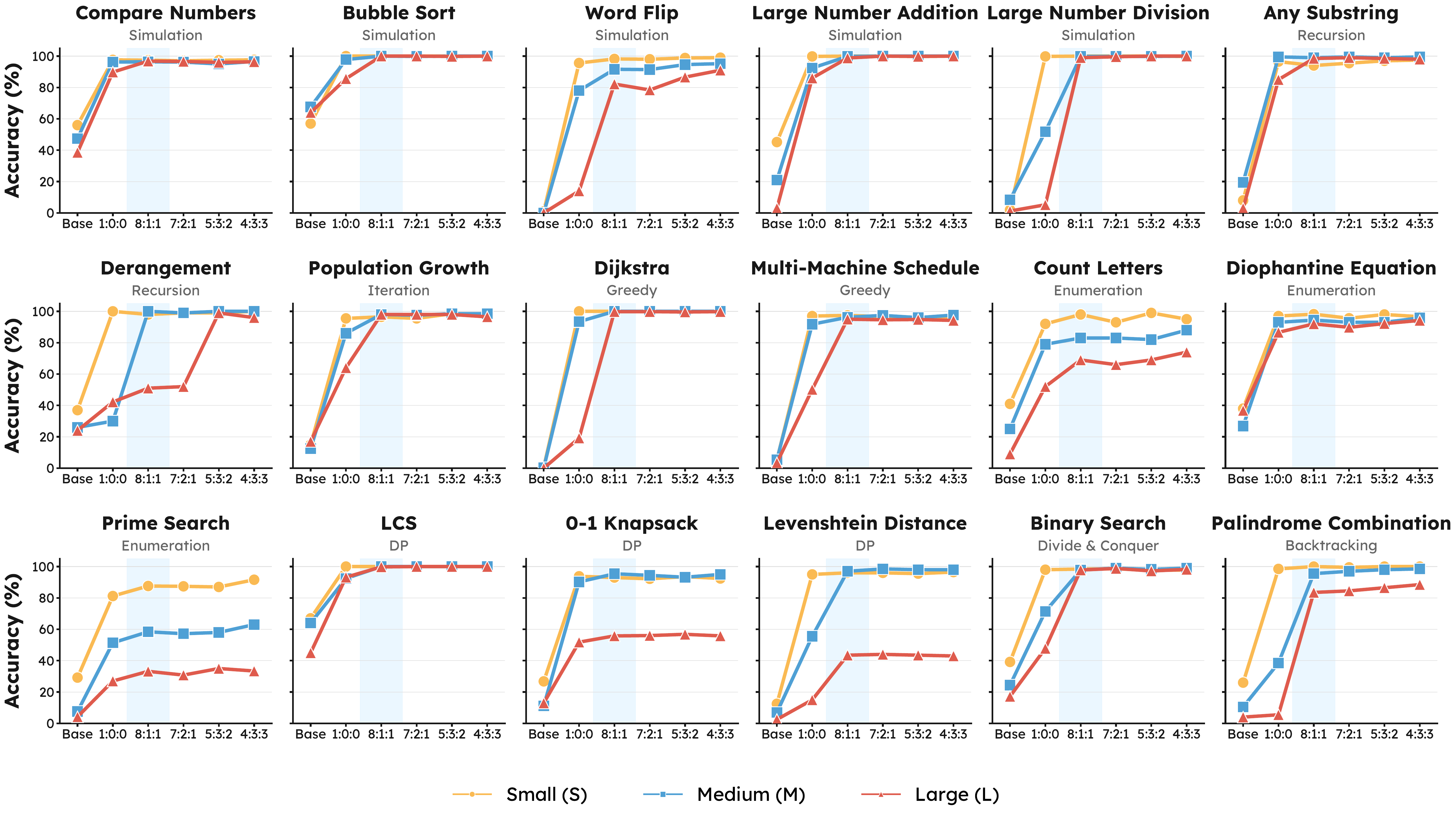}
    \vspace{-15pt}
    \caption{Experiments on the proportion of training data of different lengths. We find that adding a small amount of long data to most unsaturated tasks achieves fast performance gains.} 
    \vspace{-10pt}
    \label{fig:res_proportion}
\end{figure}

% \noindent\textbf{Observations on Data Proportions.}
\noindent\textbf{Length Generalization Activation.}
In previous experiments, we only trained on S-range data to evaluate the generalization performance on longer sequences (M, L). For tasks that haven't achieved saturation, we gradually introduce longer examples into the training set and analyze the optimal proportion for effective length generalization at minimal token cost. 
We explored five training configurations of \verb|<S,M,L>| data while keeping the total number of samples constant: \verb|<1:0:0>| (the previous method using only short sequences), \verb|<8:1:1>|, \verb|<7:2:1>|, \verb|<5:3:2>|, and \verb|<4:3:3>|. As shown in Figure~\ref{fig:res_proportion}, for almost all tasks, even a small addition of longer sequence data (\ie, at \verb|<8:1:1>|) led to a rapid saturation in long-sequence reasoning, a phenomenon we refer to as \textit{length generalization activation}.
This observation is quite different from the "balanced length" conclusion of training data in previous works~\citep{lee2023teaching}, indicating that TAIL has the potential to expand to much longer sequences at a lower cost in the future. 
See more details in Appendix~\ref{sec:proportion}.

\subsection{Ablation Study}
\label{subsec:ablation}

% \noindent{\textbf{Impact of each core module of TAIL.}}
% To verify the indispensability of each module, we removed different modules one by one and observed the performance changes.
% As shown in Table~\ref{tab:module_ablation}, the absence of any module leads to a sharp decline in length generalization performance. Moreover, we observe that the importance of each module varies across tasks. For instance, the absence of Data Review causes the most severe degradation in the Population Growth task (Iteration-based algorithm), whereas it appears to have little impact on the Compare Number task (Simulation-based algorithm).

% We hypothesize that this discrepancy arises from varying structural requirements across different task implementations. Some tasks, such as Compare Numbers, rely only on local state transitions at each step, making them less constrained by long-range dependencies and, consequently, less dependent on Data Review. Conversely, tasks with complex logical structures, such as those involving recursion-based algorithms, appear to benefit more from Linear Expansion. Additionally, most tasks are influenced by problem scale, suggesting that Atomic Subtask decomposition is a common necessity for reasoning tasks. Regardless of task type, \methodname{} integrates all three core modules, complementing each other to bridge structural gaps across different reasoning processes.

\noindent{\textbf{Key module ablation.}}
To assess the necessity of each core module, we ablate them individually and examine the performance drop. As shown in Table~\ref{tab:module_ablation}, removing any module leads to a notable decline in length generalization. Importantly, the impact varies by task: for example, \moduleTape{} is critical for Population Growth (iteration-based), but less so for Compare Numbers (simulation-based). We attribute this variation to differences in task structure. Tasks like Compare Numbers involve only local transitions and weak long-range dependencies, making \moduleTape{} less essential. In contrast, recursion-heavy tasks benefit significantly from \moduleExpan{}. Most tasks are also sensitive to scale, highlighting the general need for \moduleState{} decomposition. Overall, \methodname{} integrates all three modules synergistically to support diverse reasoning structures.

\noindent{\textbf{Thinking style ablation.}}
To investigate the influence of different CoT styles on performance, we conducted fine-tuning experiments using both standard \methodname{}-CoT and \methodname{}-CoT-\textit{styled} data.
As illustrated in Figure~\ref{fig:method_res}, the results indicate that the choice of CoT style has minimal impact on the final performance. This suggests that for the length generalization task, the specific style of CoT is not a critical factor. Instead, key modules of \methodname{} appears to play a more significant role in determining the overall performance.

\noindent{\textbf{Attention visualization of \moduleTape{}.}} 
% To investigate the impact of \moduleTape{} on attention mechanisms, we visualized the attention patterns of the model both with and without \moduleTape{}.
As shown in Figure~\ref{fig:res_attention}, when \moduleTape{} is present, we observe strong and focused attention on the corresponding tokens (highlighted in selected Transformer layers). In contrast, the attention patterns become sparse and disorganized without \moduleTape{}, showing insufficient focus on the operands. See more details in Appendix~\ref{sec:atten_vis}.

\subsection{Comparison with Reasoning Models}
\label{exp_reasoning_cmp}
It seems that both \methodname{}-CoT(-\textit{styled}) and reasoning models (\ie, DeepSeek-R1) can improve performance by extending the CoT length, but underlying principles are quite different. 
Reasoning models aim to expand the search space by prolonging the reasoning trajectory, encouraging \textbf{broad method exploration} instead of delving into the problem step by step (as shown in Section~\ref{res_dsr1_cot}). In contrast, \methodname{} focuses on \textbf{controllable} and \textbf{structured} reasoning chains that support stable generalization to longer sequences (as shown in Section~\ref{res_tail}). See Appendix~\ref{sec:cot_cmp} for more details.

Despite appearances suggesting that the linear expansion of small reasoning steps with explicit operands output make \methodname{}-CoT(-\textit{styled}) longer than traditional CoT, experimental results show that \methodname{}-CoT can achieve significantly higher accuracy at comparable CoT length to DeepSeek-R1 (as shown in Table~\ref{tab:tail_vs_deepseek}). It demonstrates that \methodname{}'s core modules are essential for promoting length generalization. See Appendix~\ref{sec:length_cmp} for more details.

To rigorously evaluate the difference, we also fine-tuned Qwen2.5-7B using an equal amount of correct data distilled by DeepSeek-R1. As shown in Table~\ref{tab:tail_vs_r1distill}, R1-Distilled-Qwen2.5-7B exhibits lower accuracy and generalization ability. See Appendix~\ref{sec:distill_cmp} for more details.

\section{Related Work}

\noindent\textbf{Length Generalization.}
Large language models (LLMs) often struggle to process inputs longer than those seen during training, a limitation referred to as length generalization~\citep{dubois2019location,newman2020eos,saparov2022language,anil2022exploring}.
Previous works primarily focus on model architecture enhancements and data-driven approaches to improve length generalization.
However, architectural enhancements modify the components (\eg, forward mechanism~\citep{fan2024looped}, attention mechanisms~\citep{duan2023interpolation}, position encodings~\citep{ruoss2023randomized,li2023functional,kazemnejad2023impact} and external queries~\citep{giannou2023looped}) of Transformers for specific tasks and further adaptation to be applicable to prevailing LLMs.
Data-driven approaches construct specific chain-of-thought (CoT) structures for training, such as digit-order reversal~\citep{zhou2024transformers,lee2023teaching,shen2023positional,zhou2023algorithms,martinez2023combining,mcleish2024transformers}, sequence padding~\citep{jelassi2023length}, and index hints~\citep{zhou2023algorithms,zhou2024transformers}, which are task-specific and lack universality.
Our \methodname{} focuses on universal data-driven approaches, exploring a more general and effective CoT structure, and directly adopting mainstream LLMs~\citep{bai2023qwen,yang2024qwen2,touvron2023llama,liu2024deepseek,team2023internlm,cai2024internlm2,bai2025intern} for fine-tuning, without modifying any components of the pretrained model.
Similar to ours, recent work~\citep{hou2024universal} adopts a Turing-like step-by-step tape update, but it is limited to specific positional encodings and data settings, and lacks verification across diverse tasks.

\noindent\textbf{Structured Chain-of-Thought Construction.}
Structured thinking demonstrably enhances the reasoning capabilities of LLMs~\citep{wei2022chain}. 
% Building upon this foundation, 
Prior research explored various recognition heuristics within CoT paradigm, aiming to imbue LLMs with more human-like thoughts~\citep{suzgun2024meta,zou2023generalizable,zheng2023take}. Concurrently, investigations into diverse structured data formats, including linear chains~\citep{wei2022chain}, hierarchical trees~\citep{yao2023tree}, interconnected graphs~\citep{besta2024graph}, and dynamically adapting structures~\citep{pandey2025adaptive}, which enable LLMs to search easily and improve the complex problem-solving performance.
In this paper, we introduce a novel approach to synthesizing structured CoT data by drawing inspiration from a Turing machine, which can handle inputs of arbitrary length.
This emulation offers a theoretically powerful advantage: length-generalizability, enabling the model to tackle problems of varying complexity, and broad applicability to the entire domain of \textit{computable} problems.
Notably, the search capability and different graph structures~\citep{besta2024graph,pandey2025adaptive,yao2023tree} and their targeted tasks can all be taken as instances of a Turing machine solving computable problems.

\noindent\textbf{Eliciting Chain-of-Thought.}
Efforts to guide LLMs in generating CoT rationales with specific, desired structures primarily involve prompting, supervised fine-tuning (SFT), and reinforcement learning (RL). 
Prompting-based methods, encompassing static example-based approaches~\cite{shao2023synthetic,zhou2022least,wang2023plan} and adaptive generation strategies~\cite{shum2023automatic,yao2023react,wang2024chain}, guide the model through carefully crafted prompts. However, these methods do not fundamentally enhance LLMs' inherent reasoning capabilities~\cite{wang2024advanced,rueda2025understanding}.
SFT methods~\cite{yu2025long,trung2024reft,kim2023cot,zhang2024chain} emphasize guiding single or multiple LLMs to synthesize training data with specific CoT structures, allowing the model to learn and imitate the target reasoning behavior.
RL-based methods~\cite{xiang2025towards,tang2025learning,shen2025satori} allow the model to generate its own CoT reasoning traces and optimize via a reward function, where the resulting structures emerge during training and remain partly uncontrollable~\cite{xu2025policy,liu2025evocot}.
Despite the progress of SFT and RL, a persistent challenge remains to mitigate shortcuts and logical omissions during CoT reasoning, especially when relying on model-generated rollouts~\cite{lightman2023let,zhang2024chain,wu2024mitigating,yu2025long}.
To address this, we explore using Python
programs to synthesize SFT CoT data to guarantee that the structure of CoT strictly mimics the execution process of a Turing Machine and reduce the possibility of learning shortcuts.

\section{Limitations}
% Although \methodname{} demonstrates strong length generalization in individual tasks, our experiments reveal that substantial room for improvement remains in compositional generalization (see Appendix~\ref{sec:task_gen_res}). Furthermore, this work focuses on computable problems with deterministic algorithms, leaving the exploration of nondeterministic problems as an open direction. Finally, while our training data significantly enhances the ability of open-source models to handle challenging samples, there remains a considerable performance gap compared to closed-source models. These limitations (as shown in Appendix~\ref{dis}) highlight promising avenues for our future research.
% Despite the strong length generalization in individual tasks, our experiments indicate that compositional generalization still leaves room for improvement (Appendix~\ref{sec:task_gen_res}). This work further centers on computable problems with deterministic algorithms, leaving nondeterministic cases as open directions. Finally, while \methodname{} markedly improves open-source models on challenging samples, a notable gap with closed-source models remains. These limitations (Appendix~\ref{dis}) point to promising avenues for future research.
\noindent{\textbf{Challenges in compositional generalization.}} 
Although \methodname{} improves the length generalization performance on each single task, the training of one task does not significantly improve the performance of other tasks under the same algorithm (as shown in Appendix~\ref{sec:task_gen_res}). In future work, we will take individual tasks as the entry point, to explore more diverse data composition strategies, with the goal of achieving compositional generalization.

\noindent{\textbf{Gap with close-source models.}}
In the experiment, we found that open-source models such as Qwen2.5 and DeepSeek-R1 series did not perform well on our tasks, but several closed-source models (\eg, O4-mini) were able to solve these problems well. While we acknowledge the strong performance of closed-source models, our focus is on bridging this gap solely through supervised fine-tuning of open-source models with \methodname{} data.

\noindent{\textbf{Challenges in modeling non-deterministic algorithmic tasks.}}
The scope of this work is limited to computable problems, and the core idea of simulating a Turing Machine is based on this assumption. However, for non-deterministic problems or open-ended reasoning, we cannot directly model an algorithm to solve it, which is a problem that \methodname{} cannot currently solve. We will actively explore ways to break through the boundaries of computable and fuzzy problems with structured CoT in future work.

\section{Conclusion}
% In this work, we propose Turing mAchine Imitation Learning (\methodname{}), a data-driven framework for enhancing length generalization of large language models (LLMs).
% Instead of designing task-specific CoT structures for limited tasks, \methodname{} provides general design principles to synthesize CoT data for a broader range of problems, which are \textit{computable}.
% \methodname{} formulate the process of solving computable problems in LLMs as a Turing machine execution process and introduce three core modules that facilitate length generalization: \moduleExpan{} to avoid shortcut learning, \moduleState{} to reduce difficulty of each subtask, and \moduleTape{} to ease the access of dynamic and long-range data in memory for the operation of current steps. By fine-tuning Qwen2.5-7B on a challenging dataset spanning 8 algorithmic categories and 18 tasks, \methodname{} achieve significant length generalization, surpassing DeepSeek-R1 in long-sequence reasoning.

We introduced Turing mAchine Imitation Learning (\methodname{}), a data-driven framework that instantiates three core modules (\ie, \moduleExpan{}, \moduleState{}, and \moduleTape{}) to align CoT structure with program execution and thereby promote universal and effective length generalization. Across 8 algorithm classes and 18 tasks, fine-tuning on \methodname{}-synthesized data yields strong length generalization on out-of-distribution sequence lengths, with consistent gains on difficult cases and performance that surpasses DeepSeek-R1. Unlike reasoning models that expand trajectories to explore many heuristics, TAIL enforces a controllable and step-by-step execution, which supports stable extrapolation to inputs of arbitrary length. 
% Finally, we acknowledge remaining limitations about compositional generalization in Appendix~\ref{dis}, highlighting the targets of our future work.

\clearpage
\newpage

\bibliography{main}

\begin{thebibliography}{66}
\providecommand{\natexlab}[1]{#1}
\providecommand{\url}[1]{\texttt{#1}}
\expandafter\ifx\csname urlstyle\endcsname\relax
  \providecommand{\doi}[1]{doi: #1}\else
  \providecommand{\doi}{doi: \begingroup \urlstyle{rm}\Url}\fi

\bibitem[Ahuja \& Mansouri(2024)Ahuja and Mansouri]{ahuja2024provable}
Kartik Ahuja and Amin Mansouri.
\newblock On provable length and compositional generalization.
\newblock \emph{arXiv preprint arXiv:2402.04875}, 2024.

\bibitem[Anil et~al.(2022)Anil, Wu, Andreassen, Lewkowycz, Misra, Ramasesh, Slone, Gur-Ari, Dyer, and Neyshabur]{anil2022exploring}
Cem Anil, Yuhuai Wu, Anders Andreassen, Aitor Lewkowycz, Vedant Misra, Vinay Ramasesh, Ambrose Slone, Guy Gur-Ari, Ethan Dyer, and Behnam Neyshabur.
\newblock Exploring length generalization in large language models.
\newblock \emph{Advances in Neural Information Processing Systems}, 35:\penalty0 38546--38556, 2022.

\bibitem[Arora \& Barak(2009)Arora and Barak]{arora2009computational}
Sanjeev Arora and Boaz Barak.
\newblock \emph{Computational complexity: a modern approach}.
\newblock Cambridge University Press, 2009.

\bibitem[Bai et~al.(2023)Bai, Bai, Chu, Cui, Dang, Deng, Fan, Ge, Han, Huang, et~al.]{bai2023qwen}
Jinze Bai, Shuai Bai, Yunfei Chu, Zeyu Cui, Kai Dang, Xiaodong Deng, Yang Fan, Wenbin Ge, Yu~Han, Fei Huang, et~al.
\newblock Qwen technical report.
\newblock \emph{arXiv preprint arXiv:2309.16609}, 2023.

\bibitem[Bai et~al.(2025)Bai, Cai, Cao, Cao, Chen, Chen, Chen, Chen, Chen, Chen, et~al.]{bai2025intern}
Lei Bai, Zhongrui Cai, Maosong Cao, Weihan Cao, Chiyu Chen, Haojiong Chen, Kai Chen, Pengcheng Chen, Ying Chen, Yongkang Chen, et~al.
\newblock Intern-s1: A scientific multimodal foundation model.
\newblock \emph{arXiv preprint arXiv:2508.15763}, 2025.

\bibitem[Besta et~al.(2024)Besta, Blach, Kubicek, Gerstenberger, Podstawski, Gianinazzi, Gajda, Lehmann, Niewiadomski, Nyczyk, et~al.]{besta2024graph}
Maciej Besta, Nils Blach, Ales Kubicek, Robert Gerstenberger, Michal Podstawski, Lukas Gianinazzi, Joanna Gajda, Tomasz Lehmann, Hubert Niewiadomski, Piotr Nyczyk, et~al.
\newblock Graph of thoughts: Solving elaborate problems with large language models.
\newblock In \emph{Proceedings of the AAAI Conference on Artificial Intelligence}, volume~38, pp.\  17682--17690, 2024.

\bibitem[Boolos et~al.(2002)Boolos, Burgess, and Jeffrey]{boolos2002computability}
George~S Boolos, John~P Burgess, and Richard~C Jeffrey.
\newblock \emph{Computability and logic}.
\newblock Cambridge university press, 2002.

\bibitem[Cai et~al.(2024)Cai, Cao, Chen, Chen, Chen, Chen, Chen, Chen, Chen, Chu, et~al.]{cai2024internlm2}
Zheng Cai, Maosong Cao, Haojiong Chen, Kai Chen, Keyu Chen, Xin Chen, Xun Chen, Zehui Chen, Zhi Chen, Pei Chu, et~al.
\newblock Internlm2 technical report.
\newblock \emph{arXiv preprint arXiv:2403.17297}, 2024.

\bibitem[Copeland(1997)]{copeland1997church}
B~Jack Copeland.
\newblock The church-turing thesis.
\newblock 1997.

\bibitem[Del{\'e}tang et~al.(2022)Del{\'e}tang, Ruoss, Grau-Moya, Genewein, Wenliang, Catt, Cundy, Hutter, Legg, Veness, et~al.]{deletang2022neural}
Gr{\'e}goire Del{\'e}tang, Anian Ruoss, Jordi Grau-Moya, Tim Genewein, Li~Kevin Wenliang, Elliot Catt, Chris Cundy, Marcus Hutter, Shane Legg, Joel Veness, et~al.
\newblock Neural networks and the chomsky hierarchy.
\newblock \emph{arXiv preprint arXiv:2207.02098}, 2022.

\bibitem[Duan et~al.(2023)Duan, Shi, and Xu]{duan2023interpolation}
Shaoxiong Duan, Yining Shi, and Wei Xu.
\newblock From interpolation to extrapolation: Complete length generalization for arithmetic transformers.
\newblock \emph{arXiv preprint arXiv:2310.11984}, 2023.

\bibitem[Dubois et~al.(2019)Dubois, Dagan, Hupkes, and Bruni]{dubois2019location}
Yann Dubois, Gautier Dagan, Dieuwke Hupkes, and Elia Bruni.
\newblock Location attention for extrapolation to longer sequences.
\newblock \emph{arXiv preprint arXiv:1911.03872}, 2019.

\bibitem[Fan et~al.(2024)Fan, Du, Ramchandran, and Lee]{fan2024looped}
Ying Fan, Yilun Du, Kannan Ramchandran, and Kangwook Lee.
\newblock Looped transformers for length generalization.
\newblock \emph{arXiv preprint arXiv:2409.15647}, 2024.

\bibitem[Giannou et~al.(2023)Giannou, Rajput, Sohn, Lee, Lee, and Papailiopoulos]{giannou2023looped}
Angeliki Giannou, Shashank Rajput, Jy-yong Sohn, Kangwook Lee, Jason~D Lee, and Dimitris Papailiopoulos.
\newblock Looped transformers as programmable computers.
\newblock In \emph{International Conference on Machine Learning}, pp.\  11398--11442. PMLR, 2023.

\bibitem[Guo et~al.(2025)Guo, Yang, Zhang, Song, Zhang, Xu, Zhu, Ma, Wang, Bi, et~al.]{guo2025deepseek}
Daya Guo, Dejian Yang, Haowei Zhang, Junxiao Song, Ruoyu Zhang, Runxin Xu, Qihao Zhu, Shirong Ma, Peiyi Wang, Xiao Bi, et~al.
\newblock Deepseek-r1: Incentivizing reasoning capability in llms via reinforcement learning.
\newblock \emph{arXiv preprint arXiv:2501.12948}, 2025.

\bibitem[Hopcroft et~al.(2001)Hopcroft, Motwani, and Ullman]{hopcroft2001introduction}
John~E Hopcroft, Rajeev Motwani, and Jeffrey~D Ullman.
\newblock Introduction to automata theory, languages, and computation.
\newblock \emph{Acm Sigact News}, 32\penalty0 (1):\penalty0 60--65, 2001.

\bibitem[Hou et~al.(2024)Hou, Brandfonbrener, Kakade, Jelassi, and Malach]{hou2024universal}
Kaiying Hou, David Brandfonbrener, Sham Kakade, Samy Jelassi, and Eran Malach.
\newblock Universal length generalization with turing programs.
\newblock \emph{arXiv preprint arXiv:2407.03310}, 2024.

\bibitem[Jelassi et~al.(2023)Jelassi, d'Ascoli, Domingo-Enrich, Wu, Li, and Charton]{jelassi2023length}
Samy Jelassi, St{\'e}phane d'Ascoli, Carles Domingo-Enrich, Yuhuai Wu, Yuanzhi Li, and Fran{\c{c}}ois Charton.
\newblock Length generalization in arithmetic transformers.
\newblock \emph{arXiv preprint arXiv:2306.15400}, 2023.

\bibitem[Kazemnejad et~al.(2023)Kazemnejad, Padhi, Natesan~Ramamurthy, Das, and Reddy]{kazemnejad2023impact}
Amirhossein Kazemnejad, Inkit Padhi, Karthikeyan Natesan~Ramamurthy, Payel Das, and Siva Reddy.
\newblock The impact of positional encoding on length generalization in transformers.
\newblock \emph{Advances in Neural Information Processing Systems}, 36:\penalty0 24892--24928, 2023.

\bibitem[Kim et~al.(2023)Kim, Joo, Kim, Jang, Ye, Shin, and Seo]{kim2023cot}
Seungone Kim, Se~June Joo, Doyoung Kim, Joel Jang, Seonghyeon Ye, Jamin Shin, and Minjoon Seo.
\newblock The cot collection: Improving zero-shot and few-shot learning of language models via chain-of-thought fine-tuning.
\newblock \emph{arXiv preprint arXiv:2305.14045}, 2023.

\bibitem[Lee et~al.(2023)Lee, Sreenivasan, Lee, Lee, and Papailiopoulos]{lee2023teaching}
Nayoung Lee, Kartik Sreenivasan, Jason~D Lee, Kangwook Lee, and Dimitris Papailiopoulos.
\newblock Teaching arithmetic to small transformers.
\newblock \emph{arXiv preprint arXiv:2307.03381}, 2023.

\bibitem[Li et~al.(2023)Li, You, Guruganesh, Ainslie, Ontanon, Zaheer, Sanghai, Yang, Kumar, and Bhojanapalli]{li2023functional}
Shanda Li, Chong You, Guru Guruganesh, Joshua Ainslie, Santiago Ontanon, Manzil Zaheer, Sumit Sanghai, Yiming Yang, Sanjiv Kumar, and Srinadh Bhojanapalli.
\newblock Functional interpolation for relative positions improves long context transformers.
\newblock \emph{arXiv preprint arXiv:2310.04418}, 2023.

\bibitem[Li et~al.(2024)Li, Liu, Zhou, and Ma]{li2024chain}
Zhiyuan Li, Hong Liu, Denny Zhou, and Tengyu Ma.
\newblock Chain of thought empowers transformers to solve inherently serial problems.
\newblock \emph{arXiv preprint arXiv:2402.12875}, 1, 2024.

\bibitem[Lightman et~al.(2023)Lightman, Kosaraju, Burda, Edwards, Baker, Lee, Leike, Schulman, Sutskever, and Cobbe]{lightman2023let}
Hunter Lightman, Vineet Kosaraju, Yuri Burda, Harrison Edwards, Bowen Baker, Teddy Lee, Jan Leike, John Schulman, Ilya Sutskever, and Karl Cobbe.
\newblock Let's verify step by step.
\newblock In \emph{The Twelfth International Conference on Learning Representations}, 2023.

\bibitem[Liu et~al.(2024)Liu, Feng, Xue, Wang, Wu, Lu, Zhao, Deng, Zhang, Ruan, et~al.]{liu2024deepseek}
Aixin Liu, Bei Feng, Bing Xue, Bingxuan Wang, Bochao Wu, Chengda Lu, Chenggang Zhao, Chengqi Deng, Chenyu Zhang, Chong Ruan, et~al.
\newblock Deepseek-v3 technical report.
\newblock \emph{arXiv preprint arXiv:2412.19437}, 2024.

\bibitem[Liu et~al.(2025)Liu, Li, Yu, Chen, Dong, Wang, XiaoLong, and Li]{liu2025evocot}
Huanyu Liu, Jia Li, Chang Yu, Taozhi Chen, Yihong Dong, Lecheng Wang, Hu~XiaoLong, and Ge~Li.
\newblock Evocot: Overcoming the exploration bottleneck in reinforcement learning.
\newblock \emph{arXiv preprint arXiv:2508.07809}, 2025.

\bibitem[Mart{\'\i}nez et~al.(2023)Mart{\'\i}nez, Watson, Reviriego, Hern{\'a}ndez, Juarez, and Sarkar]{martinez2023combining}
Gonzalo Mart{\'\i}nez, Lauren Watson, Pedro Reviriego, Jos{\'e}~Alberto Hern{\'a}ndez, Marc Juarez, and Rik Sarkar.
\newblock Combining generative artificial intelligence (ai) and the internet: Heading towards evolution or degradation?
\newblock \emph{arXiv preprint arXiv:2303.01255}, 2023.

\bibitem[McLeish et~al.(2024)McLeish, Bansal, Stein, Jain, Kirchenbauer, Bartoldson, Kailkhura, Bhatele, Geiping, Schwarzschild, et~al.]{mcleish2024transformers}
Sean McLeish, Arpit Bansal, Alex Stein, Neel Jain, John Kirchenbauer, Brian Bartoldson, Bhavya Kailkhura, Abhinav Bhatele, Jonas Geiping, Avi Schwarzschild, et~al.
\newblock Transformers can do arithmetic with the right embeddings.
\newblock \emph{Advances in Neural Information Processing Systems}, 37:\penalty0 108012--108041, 2024.

\bibitem[Newman et~al.(2020)Newman, Hewitt, Liang, and Manning]{newman2020eos}
Benjamin Newman, John Hewitt, Percy Liang, and Christopher~D Manning.
\newblock The eos decision and length extrapolation.
\newblock \emph{arXiv preprint arXiv:2010.07174}, 2020.

\bibitem[Pandey et~al.(2025)Pandey, Ghukasyan, Goktas, and Radha]{pandey2025adaptive}
Tushar Pandey, Ara Ghukasyan, Oktay Goktas, and Santosh~Kumar Radha.
\newblock Adaptive graph of thoughts: Test-time adaptive reasoning unifying chain, tree, and graph structures.
\newblock \emph{arXiv preprint arXiv:2502.05078}, 2025.

\bibitem[Press et~al.(2021)Press, Smith, and Lewis]{press2021train}
Ofir Press, Noah~A Smith, and Mike Lewis.
\newblock Train short, test long: Attention with linear biases enables input length extrapolation.
\newblock \emph{arXiv preprint arXiv:2108.12409}, 2021.

\bibitem[Rueda et~al.(2025)Rueda, Hassan, Perivolaris, Teferra, Samavi, Rambhatla, Wu, Zhang, Cao, Sharma, et~al.]{rueda2025understanding}
Alice Rueda, Mohammed~S Hassan, Argyrios Perivolaris, Bazen~G Teferra, Reza Samavi, Sirisha Rambhatla, Yuqi Wu, Yanbo Zhang, Bo~Cao, Divya Sharma, et~al.
\newblock Understanding llm scientific reasoning through promptings and model's explanation on the answers.
\newblock \emph{arXiv preprint arXiv:2505.01482}, 2025.

\bibitem[Ruoss et~al.(2023)Ruoss, Del{\'e}tang, Genewein, Grau-Moya, Csord{\'a}s, Bennani, Legg, and Veness]{ruoss2023randomized}
Anian Ruoss, Gr{\'e}goire Del{\'e}tang, Tim Genewein, Jordi Grau-Moya, R{\'o}bert Csord{\'a}s, Mehdi Bennani, Shane Legg, and Joel Veness.
\newblock Randomized positional encodings boost length generalization of transformers.
\newblock \emph{arXiv preprint arXiv:2305.16843}, 2023.

\bibitem[Saparov \& He(2022)Saparov and He]{saparov2022language}
Abulhair Saparov and He~He.
\newblock Language models are greedy reasoners: A systematic formal analysis of chain-of-thought.
\newblock \emph{arXiv preprint arXiv:2210.01240}, 2022.

\bibitem[Saparov et~al.(2024)Saparov, Pawar, Pimpalgaonkar, Joshi, Pang, Padmakumar, Kazemi, Kim, and He]{saparov2024transformers}
Abulhair Saparov, Srushti Pawar, Shreyas Pimpalgaonkar, Nitish Joshi, Richard~Yuanzhe Pang, Vishakh Padmakumar, Seyed~Mehran Kazemi, Najoung Kim, and He~He.
\newblock Transformers struggle to learn to search.
\newblock \emph{arXiv preprint arXiv:2412.04703}, 2024.

\bibitem[Shao et~al.(2023)Shao, Gong, Shen, Huang, Duan, and Chen]{shao2023synthetic}
Zhihong Shao, Yeyun Gong, Yelong Shen, Minlie Huang, Nan Duan, and Weizhu Chen.
\newblock Synthetic prompting: Generating chain-of-thought demonstrations for large language models.
\newblock In \emph{International Conference on Machine Learning}, pp.\  30706--30775. PMLR, 2023.

\bibitem[Shen et~al.(2025)Shen, Zeng, Qi, Hong, Chen, Lu, Wornell, Das, Cox, and Gan]{shen2025satori}
Maohao Shen, Guangtao Zeng, Zhenting Qi, Zhang-Wei Hong, Zhenfang Chen, Wei Lu, Gregory Wornell, Subhro Das, David Cox, and Chuang Gan.
\newblock Satori: Reinforcement learning with chain-of-action-thought enhances llm reasoning via autoregressive search.
\newblock \emph{arXiv preprint arXiv:2502.02508}, 2025.

\bibitem[Shen et~al.(2023)Shen, Bubeck, Eldan, Lee, Li, and Zhang]{shen2023positional}
Ruoqi Shen, S{\'e}bastien Bubeck, Ronen Eldan, Yin~Tat Lee, Yuanzhi Li, and Yi~Zhang.
\newblock Positional description matters for transformers arithmetic.
\newblock \emph{arXiv preprint arXiv:2311.14737}, 2023.

\bibitem[Shi et~al.(2022)Shi, Hong, Zaheer, Yin, and Sutton]{shi2022compositional}
Kensen Shi, Joey Hong, Manzil Zaheer, Pengcheng Yin, and Charles Sutton.
\newblock Compositional generalization and decomposition in neural program synthesis.
\newblock \emph{arXiv preprint arXiv:2204.03758}, 2022.

\bibitem[Shum et~al.(2023)Shum, Diao, and Zhang]{shum2023automatic}
KaShun Shum, Shizhe Diao, and Tong Zhang.
\newblock Automatic prompt augmentation and selection with chain-of-thought from labeled data.
\newblock \emph{arXiv preprint arXiv:2302.12822}, 2023.

\bibitem[Sinha et~al.(2024)Sinha, Premsri, and Kordjamshidi]{sinha2024survey}
Sania Sinha, Tanawan Premsri, and Parisa Kordjamshidi.
\newblock A survey on compositional learning of ai models: Theoretical and experimetnal practices.
\newblock \emph{arXiv preprint arXiv:2406.08787}, 2024.

\bibitem[Sipser(1996)]{sipser1996introduction}
Michael Sipser.
\newblock Introduction to the theory of computation.
\newblock \emph{ACM Sigact News}, 27\penalty0 (1):\penalty0 27--29, 1996.

\bibitem[Suzgun \& Kalai(2024)Suzgun and Kalai]{suzgun2024meta}
Mirac Suzgun and Adam~Tauman Kalai.
\newblock Meta-prompting: Enhancing language models with task-agnostic scaffolding.
\newblock \emph{arXiv preprint arXiv:2401.12954}, 2024.

\bibitem[Tang et~al.(2025)Tang, Wang, and Munos]{tang2025learning}
Yunhao Tang, Sid Wang, and R{\'e}mi Munos.
\newblock Learning to chain-of-thought with jensen's evidence lower bound.
\newblock \emph{arXiv preprint arXiv:2503.19618}, 2025.

\bibitem[Team(2023)]{team2023internlm}
InternLM Team.
\newblock Internlm: A multilingual language model with progressively enhanced capabilities, 2023.

\bibitem[Touvron et~al.(2023)Touvron, Martin, Stone, Albert, Almahairi, Babaei, Bashlykov, Batra, Bhargava, Bhosale, et~al.]{touvron2023llama}
Hugo Touvron, Louis Martin, Kevin Stone, Peter Albert, Amjad Almahairi, Yasmine Babaei, Nikolay Bashlykov, Soumya Batra, Prajjwal Bhargava, Shruti Bhosale, et~al.
\newblock Llama 2: Open foundation and fine-tuned chat models.
\newblock \emph{arXiv preprint arXiv:2307.09288}, 2023.

\bibitem[Trung et~al.(2024)Trung, Zhang, Jie, Sun, Jin, and Li]{trung2024reft}
Luong Trung, Xinbo Zhang, Zhanming Jie, Peng Sun, Xiaoran Jin, and Hang Li.
\newblock Reft: Reasoning with reinforced fine-tuning.
\newblock In \emph{Proceedings of the 62nd Annual Meeting of the Association for Computational Linguistics (Volume 1: Long Papers)}, pp.\  7601--7614, 2024.

\bibitem[Turing et~al.(1936)]{turing1936computable}
Alan~Mathison Turing et~al.
\newblock On computable numbers, with an application to the entscheidungsproblem.
\newblock \emph{J. of Math}, 58\penalty0 (345-363):\penalty0 5, 1936.

\bibitem[Wang et~al.(2024)Wang, Sun, Gong, Ye, Chen, Zhao, Liang, and Hao]{wang2024advanced}
Guoqing Wang, Zeyu Sun, Zhihao Gong, Sixiang Ye, Yizhou Chen, Yifan Zhao, Qingyuan Liang, and Dan Hao.
\newblock Do advanced language models eliminate the need for prompt engineering in software engineering?
\newblock \emph{arXiv preprint arXiv:2411.02093}, 2024.

\bibitem[Wang et~al.(2023)Wang, Xu, Lan, Hu, Lan, Lee, and Lim]{wang2023plan}
Lei Wang, Wanyu Xu, Yihuai Lan, Zhiqiang Hu, Yunshi Lan, Roy Ka-Wei Lee, and Ee-Peng Lim.
\newblock Plan-and-solve prompting: Improving zero-shot chain-of-thought reasoning by large language models.
\newblock \emph{arXiv preprint arXiv:2305.04091}, 2023.

\bibitem[Wang et~al.(2025)Wang, Huang, Song, Zhang, Iwasawa, Matsuo, and Guo]{wang2025beyond}
Ru~Wang, Wei Huang, Selena Song, Haoyu Zhang, Yusuke Iwasawa, Yutaka Matsuo, and Jiaxian Guo.
\newblock Beyond in-distribution success: Scaling curves of cot granularity for language model generalization.
\newblock \emph{arXiv preprint arXiv:2502.18273}, 2025.

\bibitem[Wang \& Zhou(2024)Wang and Zhou]{wang2024chain}
Xuezhi Wang and Denny Zhou.
\newblock Chain-of-thought reasoning without prompting.
\newblock \emph{arXiv preprint arXiv:2402.10200}, 2024.

\bibitem[Wei et~al.(2022)Wei, Wang, Schuurmans, Bosma, Xia, Chi, Le, Zhou, et~al.]{wei2022chain}
Jason Wei, Xuezhi Wang, Dale Schuurmans, Maarten Bosma, Fei Xia, Ed~Chi, Quoc~V Le, Denny Zhou, et~al.
\newblock Chain-of-thought prompting elicits reasoning in large language models.
\newblock \emph{Advances in neural information processing systems}, 35:\penalty0 24824--24837, 2022.

\bibitem[Wu et~al.(2024)Wu, Zhang, and Zhao]{wu2024mitigating}
Yexin Wu, Zhuosheng Zhang, and Hai Zhao.
\newblock Mitigating misleading chain-of-thought reasoning with selective filtering.
\newblock \emph{arXiv preprint arXiv:2403.19167}, 2024.

\bibitem[Xiang et~al.(2025)Xiang, Snell, Gandhi, Albalak, Singh, Blagden, Phung, Rafailov, Lile, Mahan, et~al.]{xiang2025towards}
Violet Xiang, Charlie Snell, Kanishk Gandhi, Alon Albalak, Anikait Singh, Chase Blagden, Duy Phung, Rafael Rafailov, Nathan Lile, Dakota Mahan, et~al.
\newblock Towards system 2 reasoning in llms: Learning how to think with meta chain-of-though.
\newblock \emph{arXiv preprint arXiv:2501.04682}, 2025.

\bibitem[Xu(2025)]{xu2025policy}
Xingcheng Xu.
\newblock The policy cliff: A theoretical analysis of reward-policy maps in large language models.
\newblock \emph{arXiv preprint arXiv:2507.20150}, 2025.

\bibitem[Yang et~al.(2024)Yang, Yang, Zhang, Hui, Zheng, Yu, Li, Liu, Huang, Wei, et~al.]{yang2024qwen2}
An~Yang, Baosong Yang, Beichen Zhang, Binyuan Hui, Bo~Zheng, Bowen Yu, Chengyuan Li, Dayiheng Liu, Fei Huang, Haoran Wei, et~al.
\newblock Qwen2. 5 technical report.
\newblock \emph{arXiv preprint arXiv:2412.15115}, 2024.

\bibitem[Yao et~al.(2023{\natexlab{a}})Yao, Yu, Zhao, Shafran, Griffiths, Cao, and Narasimhan]{yao2023tree}
Shunyu Yao, Dian Yu, Jeffrey Zhao, Izhak Shafran, Tom Griffiths, Yuan Cao, and Karthik Narasimhan.
\newblock Tree of thoughts: Deliberate problem solving with large language models.
\newblock \emph{Advances in neural information processing systems}, 36:\penalty0 11809--11822, 2023{\natexlab{a}}.

\bibitem[Yao et~al.(2023{\natexlab{b}})Yao, Zhao, Yu, Du, Shafran, Narasimhan, and Cao]{yao2023react}
Shunyu Yao, Jeffrey Zhao, Dian Yu, Nan Du, Izhak Shafran, Karthik Narasimhan, and Yuan Cao.
\newblock React: Synergizing reasoning and acting in language models.
\newblock In \emph{International Conference on Learning Representations (ICLR)}, 2023{\natexlab{b}}.

\bibitem[Yu et~al.(2025)Yu, Yuan, Wei, Wang, Qi, and Chen]{yu2025long}
Bin Yu, Hang Yuan, Yuliang Wei, Bailing Wang, Weizhen Qi, and Kai Chen.
\newblock Long-short chain-of-thought mixture supervised fine-tuning eliciting efficient reasoning in large language models.
\newblock \emph{arXiv preprint arXiv:2505.03469}, 2025.

\bibitem[Zhang et~al.(2024)Zhang, Du, Pang, Liu, Gao, and Lin]{zhang2024chain}
Xuan Zhang, Chao Du, Tianyu Pang, Qian Liu, Wei Gao, and Min Lin.
\newblock Chain of preference optimization: Improving chain-of-thought reasoning in llms.
\newblock \emph{Advances in Neural Information Processing Systems}, 37:\penalty0 333--356, 2024.

\bibitem[Zheng et~al.(2023)Zheng, Mishra, Chen, Cheng, Chi, Le, and Zhou]{zheng2023take}
Huaixiu~Steven Zheng, Swaroop Mishra, Xinyun Chen, Heng-Tze Cheng, Ed~H Chi, Quoc~V Le, and Denny Zhou.
\newblock Take a step back: Evoking reasoning via abstraction in large language models.
\newblock \emph{arXiv preprint arXiv:2310.06117}, 2023.

\bibitem[Zhou et~al.(2022)Zhou, Sch{\"a}rli, Hou, Wei, Scales, Wang, Schuurmans, Cui, Bousquet, Le, et~al.]{zhou2022least}
Denny Zhou, Nathanael Sch{\"a}rli, Le~Hou, Jason Wei, Nathan Scales, Xuezhi Wang, Dale Schuurmans, Claire Cui, Olivier Bousquet, Quoc Le, et~al.
\newblock Least-to-most prompting enables complex reasoning in large language models.
\newblock \emph{arXiv preprint arXiv:2205.10625}, 2022.

\bibitem[Zhou et~al.(2023)Zhou, Bradley, Littwin, Razin, Saremi, Susskind, Bengio, and Nakkiran]{zhou2023algorithms}
Hattie Zhou, Arwen Bradley, Etai Littwin, Noam Razin, Omid Saremi, Josh Susskind, Samy Bengio, and Preetum Nakkiran.
\newblock What algorithms can transformers learn? a study in length generalization.
\newblock \emph{arXiv preprint arXiv:2310.16028}, 2023.

\bibitem[Zhou et~al.(2024)Zhou, Alon, Chen, Wang, Agarwal, and Zhou]{zhou2024transformers}
Yongchao Zhou, Uri Alon, Xinyun Chen, Xuezhi Wang, Rishabh Agarwal, and Denny Zhou.
\newblock Transformers can achieve length generalization but not robustly.
\newblock \emph{arXiv preprint arXiv:2402.09371}, 2024.

\bibitem[Zou et~al.(2023)Zou, Zhang, Zhao, and Tang]{zou2023generalizable}
Anni Zou, Zhuosheng Zhang, Hai Zhao, and Xiangru Tang.
\newblock Generalizable chain-of-thought prompting in mixed-task scenarios with large language models.
\newblock \emph{arXiv preprint arXiv:2310.06692}, 2023.

\end{thebibliography}
\bibliographystyle{iclr2026_conference}

% \appendix
% \section{Appendix}

\clearpage

\appendix
\setcounter{table}{0} 
\setcounter{figure}{0}
\setcounter{equation}{0}
\renewcommand{\thetable}{A\arabic{table}}
\renewcommand\thefigure{A\arabic{figure}} 
\renewcommand\theequation{A\arabic{equation}}

\section{Formal Definition of Turing Machine}
\label{sec:def_TM}

A Turing machine~\citep{turing1936computable,hopcroft2001introduction} can be formally defined as a seven-tuple:

\begin{align}
\label{eq:Pre_7tuple}
    M = (Q, \Sigma, \Gamma, \delta, q_0,B,F),
\end{align}
where $Q$ is a finite set of states, $\Sigma$ is a non-empty finite input alphabet, $\Gamma$ denotes the set of tape symbols, $\delta$ is the transition function, $q_0 \in Q$ refers to the initial state, $B \in \Gamma - \Sigma$ is the blank symbol, and $F$ denotes the set of final states.

\section{The Underlying Motivation of \methodname{}}
\label{sec:motivation}

\begin{tldrnbox}
\textbf{TL;DR.}
To improve length generalization, TAIL resolves two fundamental limitations of traditional CoT: \textbf{shortcut reasoning} and \textbf{sparse long-distance attention}. The \textit{Atomic State} and \textit{Linear Transition} address shortcut behavior at the micro and macro levels respectively, while the \textit{Memory Fetcher} mitigates the sparse-attention issue.
\end{tldrnbox}

\subsection{Shortcut Reasoning}
Previous works~\cite{anil2022exploring,saparov2024transformers} show that when SFT'd on non-sequential CoT, LLMs prefer to take shortcuts during inference, and subsequently fail to generalize to longer sequences. From a micro perspective, shortcuts appear as \textbf{overly large single-step reasoning}, which means the model jumps to substantial sub-results instead of reasoning step by step~\cite{anil2022exploring,saparov2024transformers}; From a macro perspective, we also observe shortcut behavior as \textbf{step-skipping} and \textbf{frequent switching between reasoning strategies} (see Appendix~\ref{res_dsr1_cot} for CoT example of DeepSeek-R1).

\begin{itemize}
  \item \textbf{Atomic State} focuses on the micro level of each reasoning step. Guided by the RASP-L theory~\cite{zhou2023algorithms}, it produces length-generalizable steps of appropriate granularity (\ie, steps without internal loops).
  \item \textbf{Linear Transition} focuses on the macro structure of connecting reasoning steps. It linearly unfolds the entire reasoning process step by step, and the tight logical linkage between Atomic States prevents step-skipping.
\end{itemize}

\subsection{Sparse Long-distance Attention}
Previous work~\cite{wang2025beyond} shows that long-sequence reasoning leads to severe attention sparsity in later steps. As illustrated in Figure~\ref{fig:res_attention} (lower part), traditional CoT attends to distant past states while performing the current reasoning action, resulting in sparse long-distance attention. Memory Fetcher decouples long-range attention construction from action execution: (1) First build long-range attention via a simple copy operation, copying the relevant operands to the end of the sequence; (2) then performs the reasoning action.

As shown in Fig~\ref{fig:res_attention} (upper part), introducing Memory Fetcher significantly improves attention distribution during reasoning actions. More attention is focused on nearby relevant operands, improving single-step correctness, and thereby stabilizing long-sequence reasoning.

\clearpage

\setcounter{table}{0} 
\setcounter{figure}{0}
\setcounter{equation}{0}
\renewcommand{\thetable}{C\arabic{table}}
\renewcommand\thefigure{C\arabic{figure}} 
\renewcommand\theequation{C\arabic{equation}}

\section{Task Introduction}
\label{sec:dataset}

\definecolor{brightgreen}{RGB}{50, 205, 50}
\definecolor{skyblue}{RGB}{135, 206, 235}
\definecolor{coral}{RGB}{255, 127, 80}

The dataset consists of purely synthetic data, covering 8 major algorithms and 18 tasks, as shown in Table~\ref{tab:algorithm_task_length}. Most tasks have approximately 100,000 training samples and 500 test samples (a small subset of tasks, which are more difficult to construct, retain 20,000 training samples and 200 test samples). All test queries have been verified and are not included in the training set.

\begin{table}[!th]
\centering
\renewcommand{\arraystretch}{1.2} % 调整行距
\resizebox{\textwidth}{!}{ % 让表格自动撑满页宽
\begin{tabular}{m{2.8cm}|m{4cm}|m{8cm}} % 左侧两列窄一点，右侧宽一点
\toprule[1pt]
\textbf{Algorithm} & \textbf{Task Name} & \textbf{Task Content} \\
\midrule
\multirow{5}{*}{Simulation} & Large Number Addition & \raggedright $x_1+x_2$ \textcolor[RGB]{227,23,13}{$^*$} \textcolor[RGB]{128,138,135}{($\text{len}(x_1)=n,\text{len}(x_2)=m$)}\arraybackslash \\
                              % & Large Number Product & \raggedright $x_1\times x_2$ \textcolor[RGB]{227,23,13}{$^*$} \textcolor[RGB]{128,138,135}{($\text{len}(x_1)=n,\text{len}(x_2)=m$)}\arraybackslash \\
                              % & Large Number Subtraction & \raggedright $x_1-x_2$ \textcolor[RGB]{227,23,13}{$^*$} \textcolor[RGB]{128,138,135}{($\text{len}(x_1)=n,\text{len}(x_2)=m$)}\arraybackslash \\
                              & Large Number Division & \raggedright $x_1\div x_2$ \textcolor[RGB]{227,23,13}{$^*$} \textcolor[RGB]{128,138,135}{($\text{len}(x_1)=n,\text{len}(x_2)=m$)}\arraybackslash \\
                              & Bubble Sort & \raggedright Bubble sort list of $n$ non-repeat numbers \arraybackslash \\
                              & Word Flip & \raggedright Flip a sentence containing $n$ letters \arraybackslash \\
                              & Compare Numbers & \raggedright Compare $x_1$ and $x_2$ \textcolor[RGB]{227,23,13}{$^*$} \textcolor[RGB]{128,138,135}{($\text{len}(x_1)=n,\text{len}(x_2)=m$)}\arraybackslash \\
\midrule
\multirow{2}{*}{Recursion} & Any Substring & \raggedright Find all substrings of given string with length $n$  \arraybackslash \\
                                   & Derangement & \raggedright Derangement count for $n$ elements \arraybackslash \\
                                   % & Graph DFS & \raggedright DFS path in graph containing $n$ nodes \arraybackslash \\
                                   
\midrule
\multirow{1}{*}{Iteration} 
                               & Population Growth & \raggedright 
                               Calculate total pairs after $n$ units, starting reproduction at $x$-th unit ($x < n$) with initial $y$ pairs, $z$ pairs produced per unit ($y,z \in \mathbb{N}^+$)\arraybackslash \\

\midrule
\multirow{2}{*}{Greedy} & Dijkstra & \raggedright Shortest path values in graph with $n$ vertices\arraybackslash \\
                                  % & Fractional Knapsack & \raggedright Maximum benefit of $n$-element fractional knapsack\arraybackslash \\
                                  % & Huffman & \raggedright Huffman encoding of character in string of length $n$\arraybackslash \\
                                  % & MST & \raggedright Minimal total weight of $n$-node graph's MST\arraybackslash \\
                                  & Multi-Machine Schedule & \raggedright Maximum benefit of $n$ tasks on $x$ queues ($x<n$)\arraybackslash \\
\midrule
\multirow{3}{*}{Enumeration} & Count Letters & \raggedright Count letters in sentence of length $n$\arraybackslash \\
                                      & Diophantine Equation & \raggedright Find integer solutions to $x_1 a + x_2 b = n$ ($a, b \geq 0$) \arraybackslash \\
                                      % & Prime Factorization & \raggedright All prime factors of $x$ \textcolor[RGB]{128,138,135}{($\text{len}(x)=n$)} \arraybackslash \\
                                      & Prime Search & \raggedright All prime numbers in the interval $[n,m]$ ($n<m$) \arraybackslash \\
\midrule
\multirow{3}{*}{DP} & 0-1 Knapsack & \raggedright Maximum benefit of $n$-element 0-1 knapsack \arraybackslash \\
                                     & LCS & \raggedright LCS of string $X_1$ and $X_2$ \textcolor[RGB]{128,138,135}{($\text{len}(X_1)=n,\text{len}(X_2)=m$)} \arraybackslash \\
                                     & Levenshtein Distance & \raggedright Minimum operations converting string $X_1$ to $X_2$ \arraybackslash \\
\midrule
\multirow{1}{*}{Divide \& Conquer} & Binary Search & \raggedright Binary search index in list of $n$ increasing numbers \arraybackslash \\
                                     % & Merge Sort & \raggedright Merge sort list of $n$ non-repeat numbers \arraybackslash \\
\midrule
\multirow{1}{*}{Backtracking} & Permutation Combination & \raggedright Number of combinations in $n$-element list (step by step) \arraybackslash \\
\bottomrule[1pt]
\end{tabular}
}
\vspace{1em}
\caption{Dataset synthesised under instruction of \methodname{}, containing 8 algorithms and 18 tasks. $n$ and $m$ represent length in a given range $G\in \{S,M,L\}$. \textcolor[RGB]{227,23,13}{$*$} indicates that a decimal point can be inserted in any bit of the operand in specific task. }
\label{tab:algorithm_task_length}
\end{table}

\clearpage

\setcounter{table}{0} 
\setcounter{figure}{0}
\setcounter{equation}{0}
\renewcommand{\thetable}{D\arabic{table}}
\renewcommand\thefigure{D\arabic{figure}} 
\renewcommand\theequation{D\arabic{equation}}

\section{Data Synthesis and CoT Examples}
\label{sec:Data_Flow}

% \setcounter{table}{0} 
% \setcounter{figure}{0}
% \setcounter{equation}{0}
% \renewcommand{\thetable}{B\arabic{table}}
% \renewcommand\thefigure{B\arabic{figure}} 
% \renewcommand\theequation{B\arabic{equation}}

% \section{Chain-of-Thought Format Examples}
% \label{sec:CoT_sample}

\begin{figure}[h]
    \centering
    \includegraphics[width=1\linewidth]{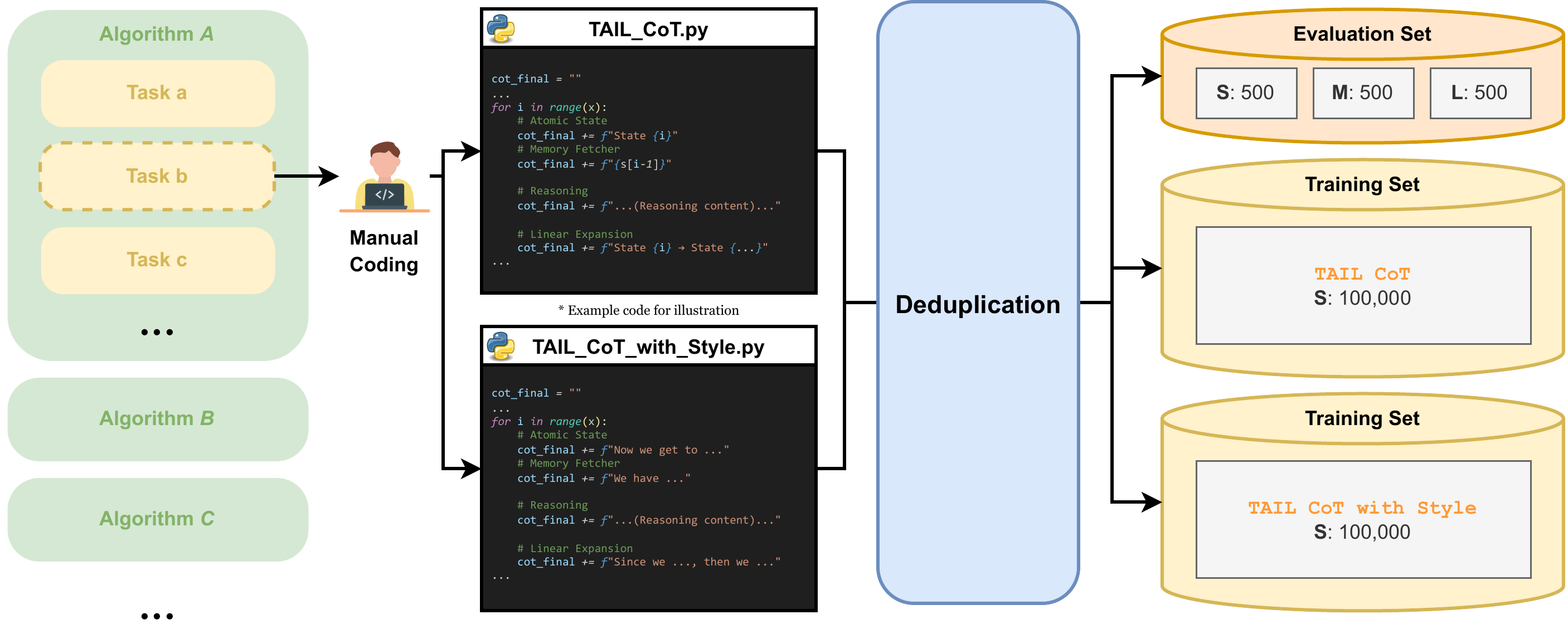}
    \caption{Overall pipeline of data synthesis. We manually code task-specific python programs and use them to massively generate chain-of-thought (CoT) data, either in plain form (\methodname{}-CoT) or with stylistic variations (\methodname{}-CoT-\textit{styled}). After deduplication, we construct large-scale training sets and balanced evaluation sets for subsequent experiments. (\textbf{S} = Short sequence data, \textbf{M} = Medium sequence data, \textbf{L} = Long sequence data)} 
    \vspace{-10pt}
    \label{fig:data_flow}
\end{figure}

Figure~\ref{fig:data_flow} illustrates our data synthesis process. Specifically, for each task, we manually write a Python program that can accept any input under that task. Because each task is assigned to an algorithm, writing this program is very convenient and well-reasoned. We then add string concatenation statements to the program to link the reasoning process and form a complete Chain-of-Thought (CoT). Since CoTs are generated as the program executes, they completely follow the program's execution process, which is the core idea behind \methodname{}. As the synthetic CoTs strictly follows the running process of Python programs, they will \textit{exhaustively explore all possible solutions}.

During this process, we can output two types of CoT: (1) \textbf{\methodname{}-CoT} contains only the TAIL core module, without any other verbiage and is more symbolic. (2) \textbf{\methodname{}-CoT-\textit{styled}} adds more style statements and is more human-readable and interpretable, which adds more cohesive and planning statements.

In experiments, we trained \methodname{}-CoT-\textit{styled} on 18 tasks across all 8 algorithm classes and verified its strong length generalization performance, as shown in Figure~\ref{fig:res_dsr1}. We then verified that removing all explicit style statements (\methodname{}-CoT) did not lead to a performance degradation, as shown in Figure~\ref{fig:method_res}, demonstrating that length generalization is the core module of TAIL, not the style statement.

Take \textit{Binary Search} task in \textit{Divide \& Conquer} algorithm as an example.
Figure~\ref{fig:query} is the query as direct input to LLMs.
For each task, we constructed more than 20 query templates to simulate the diversity.
Figure~\ref{fig:mini_cot} is an example of \methodname{}-CoT and Figure~\ref{fig:user_cot} is an example of \methodname{}-CoT-\textit{styled}.

\begin{figure*}[h] 
\begin{AIbox}{}
{\bf Query: (Example 1)} \\
{
\\
Determine the index of \$-5259\$ in the sorted list \$[-5957, -5259, -4195, -2263, 1289, 3514, 3632, 4284, 5991, 6578, 7333]\$ with binary search (start from \$0\$). Other search methods are not allowed.
\\

}
{\bf Query: (Example 2)} \\
{
\\
Find the index of \$-5259\$ in the sorted list \$-5957, -5259, -4195, -2263, 1289, 3514, 3632, 4284, 5991, 6578, 7333\$, starting from \$0\$. For teaching purposes, you must use binary search and show the process step by step.
\\

}
{\bf Query: (Example 3)} \\
{
\\
Provide the 0-based binary search index for \$-5259\$ in \$[-5957, -5259, -4195, -2263, 1289, 3514, 3632, 4284, 5991, 6578, 7333]\$.
\\
}
\end{AIbox} 
\caption{Example queries belonging to the Binary Search task (Divide \& Conquer algorithm).}
\label{fig:query}
\end{figure*}

\begin{figure*}[h] 
\begin{AIbox}{}
{\bf  Minimalist reasoning chain of \methodname{}:} \\
{
\\
{\color{brightgreen}\verb|<Atomic State>|} \{0,10\} \\
{\color{blue}\verb|<Memory Fetcher>|} [($s_0=-5957$), ($s_1=-5259$), ($s_2=-4195$), ($s_3=-2263$), ($s_4=1289$), ($s_5=3514$), ($s_6=3632$), ($s_7=4284$), ($s_8=5991$), ($s_9=6578$), ($s_{10}=7333$)] \\
{\color{blue}\verb|<Memory Fetcher>|} $\text{Find} = -5259$ \\
$\text{Mid}= 5, s_5 = 3514 > \text{Find}$ \\
{\color{red}\verb|<Linear Transition>|} \{0,10\} $\to$ \{0,4\} \\

{\color{brightgreen}\verb|<Atomic Subtask>|} \{0,4\} \\
{\color{blue}\verb|<Memory Fetcher>|} [($s_0=-5957$), ($s_1=-5259$), ($s_2=-4195$), ($s_3=-2263$), ($s_4=1289$)] \\
{\color{blue}\verb|<Memory Fetcher>|} $\text{Find} = -5259$. \\
$\text{Mid} = 2, s_2 = -4195 > \text{Find}$ \\
{\color{red}\verb|<Linear Transition>|} \{0,4\} $\to$ \{0,1\} \\

{\color{brightgreen}\verb|<Atomic Subtask>|} \{0,1\} \\
{\color{blue}\verb|<Memory Fetcher>|} [($s_0=-5957$), ($s_1=-5259$)] \\
{\color{blue}\verb|<Memory Fetcher>|} $\text{Find} = -5259$ \\
$\text{Mid} = 0, s_0 = -5957 < \text{Find}$ \\
{\color{red}\verb|<Linear Transition>|} \{1,1\} $\to$ \{1,1\}. \\

{\color{brightgreen}\verb|<Atomic Subtask>|} \{1,1\} \\
{\color{blue}\verb|<Memory Fetcher>|} [($s_1=-5259$)] \\
{\color{blue}\verb|<Memory Fetcher>|} $\text{Find} = -5259$ \\
$\text{Mid} = 1, s_1 = -5259 = \text{Find}$ \\
{\color{red}\verb|<Linear Transition>|} \{1,1\} $\to$ $F$ \\

Final Answer: Index $\boxed{1}$.
\\
}
\end{AIbox} 
\caption{Example of \methodname{}-CoT, with only core modules aligned with Turing machines. $<$TAG$>$ is for demonstration purposes only and does not exist in the actual CoT content.}
\label{fig:mini_cot}
\end{figure*}

\begin{figure*}[h] 
\begin{AIbox}{}
{\bf User-friendly CoT of \methodname{}:} \\
{
\\
Let's perform the binary search step by step. \\

1. The currently processed interval is [-5957, -5259, -4195, -2263, 1289, 3514, 3632, 4284, 5991, 6578, 7333]. \\
   First we map each number to its index: \\
    - Original index $0$: $-5957$ 
    
    - Original index $1$: $-5259$
    
    - Original index $2$: $-4195$
    
    - Original index $3$: $-2263$
    
    - Original index $4$: $1289$
    
    - Original index $5$: $3514$
    
    - Original index $6$: $3632$
    
    - Original index $7$: $4284$
    
    - Original index $8$: $5991$
    
    - Original index $9$: $6578$
    
    - Original index $10$: $7333$

   The left boundary index is $0$ and the right boundary index is $10$.
   
   Median index $\lfloor (0 + 10)/2 \rfloor = 5$, the median number is $3514$.
   
   The target number $-5259$ is less than the median number $3514$, continue these operations on the left subinterval $[-5957, -5259, -4195, -2263, 1289]$.\\

2. The currently processed interval is $[-5957, -5259, -4195, -2263, 1289]$.

   First we map each number to its index:
   
    - Original index $0$: $-5957$
    
    - Original index $1$: $-5259$
    
    - Original index $2$: $-4195$
    
    - Original index $3$: $-2263$
    
    - Original index $4$: $1289$

   The left boundary index is $0$ and the right boundary index is $4$.
   
   Median index $\lfloor 0 + 4 \rfloor = 2$, the median number is $-4195$.
   
   The target number $-5259$ is less than the median number $-4195$, continue these operations on the left subinterval $[-5957, -5259]$. \\

3. The currently processed interval is $[-5957, -5259]$.

   First we map each number to its index:
   
    - Original index $0$: $-5957$
    
    - Original index $1$: $-5259$

   The left boundary index is $0$ and the right boundary index is $1$.
   
   Median index $\lfloor 0 + 1 \rfloor = 0$, the median number is $-5957$.
   
   The target number $-5259$ is greater than the median number $-5957$, continue these operations on the right subinterval $[-5259]$. \\

4. The currently processed interval is $[-5259]$.

   First we map each number to its index:
   
    - Original index $1$: $-5259$

   The left boundary index is $1$ and the right boundary index is $1$.
   
   Median index $\lfloor 1 + 1 \rfloor = 1$, the median number is $-5259$.
   
   The target number $-5259$ is equal to the median number $-5259$, the search ends.

The target number $-5259$ is located at index $\boxed{1}$.
\\
}
\end{AIbox} 
\caption{Example of \methodname{}-CoT-\textit{styled}, improving human readability on top of core modules}
\label{fig:user_cot}
\end{figure*}

\clearpage

\setcounter{table}{0} 
\setcounter{figure}{0}
\setcounter{equation}{0}
\renewcommand{\thetable}{E\arabic{table}}
\renewcommand\thefigure{E\arabic{figure}} 
\renewcommand\theequation{E\arabic{equation}}

\section{Data Proportion Study}
\label{sec:proportion}

For each task, we divided the data into three length ranges (S, M, and L), and synthesized 100,000 training samples for each range. In studying length generalization, we trained solely on the S-range data without including any M- or L-range samples (\ie, the data proportion is \verb|<1:0:0>|), and then evaluated on all three ranges to assess the ability to generalize to longer sequences. In this section, for tasks that haven't reached saturation, we progressively incorporate longer sequences into the S-range training data and investigate the data proportion that achieves saturation performance with the minimal number of training tokens. Specifically, we keep the total number of training samples fixed, while varying the proportions of the three length ranges as \verb|<1:0:0>|, \verb|<8:1:1>|, \verb|<7:2:1>|, \verb|<5:3:2>|, and \verb|<4:3:3>|.

As shown in Figure~\ref{fig:res_proportion}, we found that using \methodname{}-CoT, performance saturation can be quickly achieved by simply \textit{adding a small amount of long data} to a large amount of short data. 
Guided by this observation, \methodname{}-CoT can leverage imbalanced sequence length proportions to reduce training costs. We call this \textbf{length generalization activation}.

% \clearpage

\setcounter{table}{0} 
\setcounter{figure}{0}
\setcounter{equation}{0}
\renewcommand{\thetable}{F\arabic{table}}
\renewcommand\thefigure{F\arabic{figure}} 
\renewcommand\theequation{F\arabic{equation}}

\section{Task Length Range}
\label{sec:length_range}

\begin{table}[!th]
\centering
\renewcommand{\arraystretch}{1.2} % 调整行距
\resizebox{\textwidth}{!}{ % 让表格自动撑满页宽
\begin{tabular}{m{2.8cm}|p{4cm}|p{3cm}|p{3cm}|p{3cm}} 
\toprule[1pt]
\textbf{Algorithm} & \textbf{Task Name} & \textbf{Small (S)} & \textbf{Medium (M)} & \textbf{Long (L)}  \\
\midrule
\multirow{5}{*}{Simulation} & Large Number Addition & \raggedright $[10,30]$ & $[31,40]$ & $[41,50]$ \arraybackslash  \\
                              & Large Number Division & \raggedright $[2,5]$ & $[6,10]$ & $[11,20]$\arraybackslash \\
                              & Bubble Sort & \raggedright $[2,4]$ & $[5,6]$ & $[7,8]$ \arraybackslash  \\
                              & Word Flip & \raggedright $[10,20]$ & $[21,50]$ & $[51,100]$ \arraybackslash  \\
                              & Compare Numbers & \raggedright $[5,10]$ & $[11,20]$ & $[21,50]$ \arraybackslash \\
\midrule
\multirow{2}{*}{Recursion} & Any Substring & \raggedright $[3,5]$ & $[6,9]$ & $[10,14]$ \arraybackslash \\
                                   & Derangement & \raggedright $[3,30]$ & $[31,60]$ & $[61,100]$ \\
                                   
\midrule
\multirow{1}{*}{Iteration} & Population Growth & \raggedright $[1,10]$ & $[11,25]$ & $[26,50]$ \arraybackslash \\

\midrule
\multirow{2}{*}{Greedy} & Dijkstra & \raggedright $[3,5]$ & $[6,10]$ & $[11,20]$ \arraybackslash \\
                                  & Multi-Machine Schedule & \raggedright $[5,10]$ & $[11,20]$ & $[21,50]$ \arraybackslash \\
\midrule
\multirow{3}{*}{Enumeration} & Count Letters & \raggedright $[2,6]$ & $[7,10]$ & $[11,20]$ \arraybackslash \\
                                      & Diophantine Equation & \raggedright $[10,50]$ & $[51,100]$ & $[101,200]$ \arraybackslash \\
                                      & Prime Search & \raggedright $[5,100]$ & $[101,200]$ & $[201,300]$ \arraybackslash \\
\midrule
\multirow{3}{*}{DP} & 0-1 Knapsack & \raggedright $[2,3]$ & $[4,5]$ & $[6,8]$ \arraybackslash \\
                                     & LCS & \raggedright $[2,6]$ & $[7,9]$ & $[10,12]$ \arraybackslash \\
                                     & Levenshtein Distance & \raggedright $[2,4]$ & $[5,7]$ & $[8,9]$ \arraybackslash \\
\midrule
\multirow{1}{*}{Divide \& Conquer} & Binary Search & \raggedright $[5,20]$ & $[21,40]$ & $[41,70]$ \arraybackslash \\
\midrule
\multirow{1}{*}{Backtracking} & Palindrome Combination & \raggedright $[2,4]$ & $[5,6]$ & $[7,8]$ \arraybackslash \\
\bottomrule[1pt]
\end{tabular}
}
\vspace{1em}
\caption{The setting of length ranges across all tasks. See Table~\ref{tab:algorithm_task_length} for the definitions of \textit{length}.}
\label{tab:task_length_range}
\end{table}

%  The specific definition of \textit{length} for each task is detailed in Table~\ref{tab:algorithm_task_length}, where length range $G\in \{S,M,L\}$.

% \clearpage

\setcounter{table}{0} 
\setcounter{figure}{0}
\setcounter{equation}{0}
\renewcommand{\thetable}{G\arabic{table}}
\renewcommand\thefigure{G\arabic{figure}} 
\renewcommand\theequation{G\arabic{equation}}

\section{Baseline Data Construction}
\label{sec:baseline_data}

This section describes the data construction methods for the baseline methods (\ie, Index Hint and Reversed Format). The experimental task was Large Number Addition (Simulation algorithm), and unlike the experiments in previous works, in this paper we contain a random number of decimals. We followed data construction methods accordingly based on the principles of these baselines.

\subsection{Index Hint}
The Index Hint\citep{zhou2023algorithms,zhou2024transformers} method refers to adding a hint of indexes to the corresponding numeric or logical bits of two operands for positioning in arithmetic or parity operations. This method has been extensively proven to be effective in both tasks. To compare the performance of Index Hint and \methodname{}, we refer to the method~\citep{zhou2024transformers} that displays the indexes to locate as follows:
\begin{center}
    \verb|3|{\color{red}\verb|a|}\verb|6|{\color{red}\verb|b|}\verb|1|{\color{red}\verb|c|}\verb|+5|{\color{red}\verb|a|}\verb|7|{\color{red}\verb|b|}\verb|6|{\color{red}\verb|c|}\verb|=9|{\color{red}\verb|a|}\verb|3|{\color{red}\verb|b|}\verb|7|{\color{red}\verb|c|}
\end{center}

However, the above approach is for the case where two operands have the same number of digits without decimals, so we make following improvements for non-fixed-length cases with decimals:

\begin{center}
    \verb|1|{\color{red}\verb|(-c)|}\verb|2|{\color{red}\verb|(-b)|}\verb|3|{\color{red}\verb|(-a)|}\verb|.4|{\color{red}\verb|(a)|}\verb|5|{\color{red}\verb|(b)|}\verb|+6|{\color{red}\verb|(-b)|}\verb|7|{\color{red}\verb|(-a)|}\verb|.8|{\color{red}\verb|(a)|}\verb|9|    {\color{red}\verb|(b)|} \\ \verb|=1|{\color{red}\verb|(-c)|}\verb|9|{\color{red}\verb|(-b)|}\verb|1|{\color{red}\verb|(-a)|}\verb|.3|{\color{red}\verb|(a)|}\verb|4|{\color{red}\verb|(b)|}
\end{center}

\subsection{Reversed Format}
Reversed Format~\citep{zhou2024transformers,lee2023teaching,shen2023positional,zhou2023algorithms,martinez2023combining,mcleish2024transformers} refers to reversing each of the two operands in arithmetic operations such as addition. The rationale for this method is that the addition is usually performed from the first digit, \ie, from right to left. However, the order of next token prediction (NTP) in large language models (LLMs) is from left to right, which leads to overly complex search paths during model learning and affects the length generalization performance. This method is also a widely proven effective length generalization facilitation method, constructed as follows:

\begin{center}
    (Origin) \verb|123.45+67.89=191.34| \\
    (Reversed Format) {\color{red}\verb|54.321+98.76=43.191|}
\end{center}

% \clearpage
% \newpage

% \setcounter{table}{0} 
% \setcounter{figure}{0}
% \setcounter{equation}{0}
% \renewcommand{\thetable}{F\arabic{table}}
% \renewcommand\thefigure{F\arabic{figure}} 
% \renewcommand\theequation{F\arabic{equation}}

% \section{Task Generalization Results}
% \label{sec:task_gen_res}

% \begin{figure}[h]
%     \centering
%     \includegraphics[width=0.99\linewidth]{./figures/task_generalization.pdf}
%     \caption{Generalization performance between tasks within a single algorithm (\eg, DP and Greedy). \textbf{Pretrain} represents Qwen 2.5-7B as the basis for subsequent SFT. For each task, \textbf{SFT(Rest)} indicates training using data from other tasks within the algorithm, \textbf{SFT(Task)} indicates training using data from this task, and \textbf{SFT(All)} indicates training using data from all tasks within this algorithm. \textcolor{red}{$\blacktriangledown$} indicates that this piece of data has a label accuracy of less than 5\%.} 
%     \vspace{-10pt}
%     \label{fig:task_gen}
% \end{figure}

% \clearpage

\setcounter{table}{0} 
\setcounter{figure}{0}
\setcounter{equation}{0}
\renewcommand{\thetable}{H\arabic{table}}
\renewcommand\thefigure{H\arabic{figure}} 
\renewcommand\theequation{H\arabic{equation}}

\section{Details about \moduleTape{}}
\label{sec:atten_vis}

This section presents the details of the \moduleTape{} in \methodname{}. As shown in Figure~\ref{fig:MF_cmp}, the \moduleTape{} is designed to \textit{\textbf{decouple}} long-range attention from the reasoning action. It first retrieves all relevant operands from the long sequence to the end, and then performs more precise reasoning through local attention.

% \begin{figure}[h]
%     \centering
%     \includegraphics[width=1\linewidth]{./figures/MF_workflow.pdf}
%     \caption{Comparison of traditional CoT and \methodname{}'s Memory Fetcher, which can decouple the long-range attention construction and the reasoning process.}
%     \label{fig:MF_cmp}
% % \end{figure}

\begin{figure}[h]
    \centering
    \includegraphics[width=0.9\linewidth]{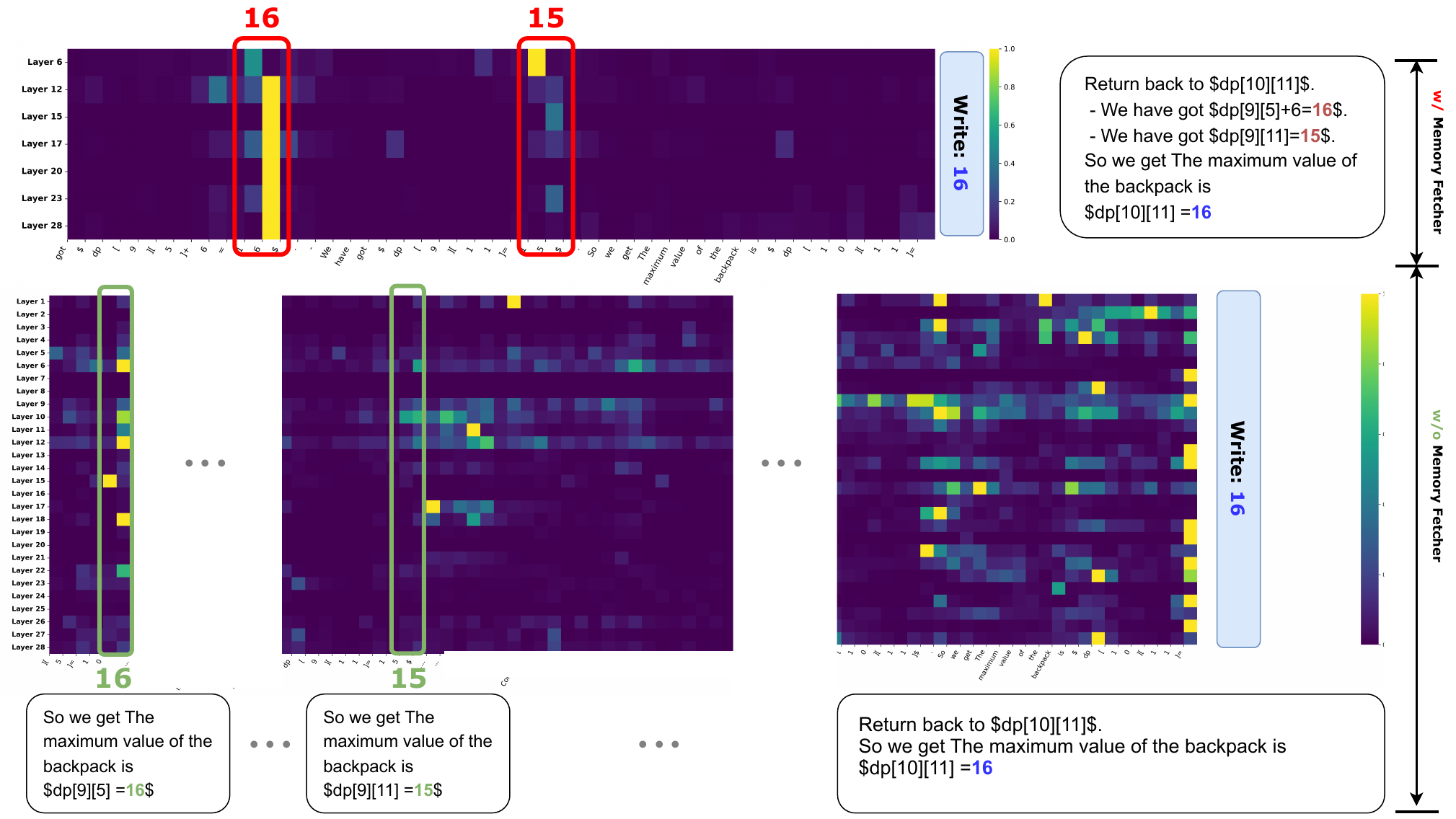}
    \caption{Ablation study on attention visualization of \moduleTape{}.}
    \label{fig:res_attention}
\end{figure}

We visualize this local attention and compare it with the traditional CoT. As shown in Figure~\ref{fig:res_attention}, attention across layers tends to focus on the end of the sequence. With the introduction of the \moduleTape{}, operands are captured more accurately. In contrast, traditional CoT must simultaneously attend to reasoning actions (via local attention) and the retrieval of distant operands (via long-range attention), which results in a significant sparsification of long-range attention.

% 任务泛化
\clearpage

\setcounter{table}{0} 
\setcounter{figure}{0}
\setcounter{equation}{0}
\renewcommand{\thetable}{I\arabic{table}}
\renewcommand\thefigure{I\arabic{figure}} 
\renewcommand\theequation{I\arabic{equation}}
\section{Combinatorial Generalization Results}
\label{sec:task_gen_res}

\begin{figure}[h]
    \centering
    \includegraphics[width=1\linewidth]{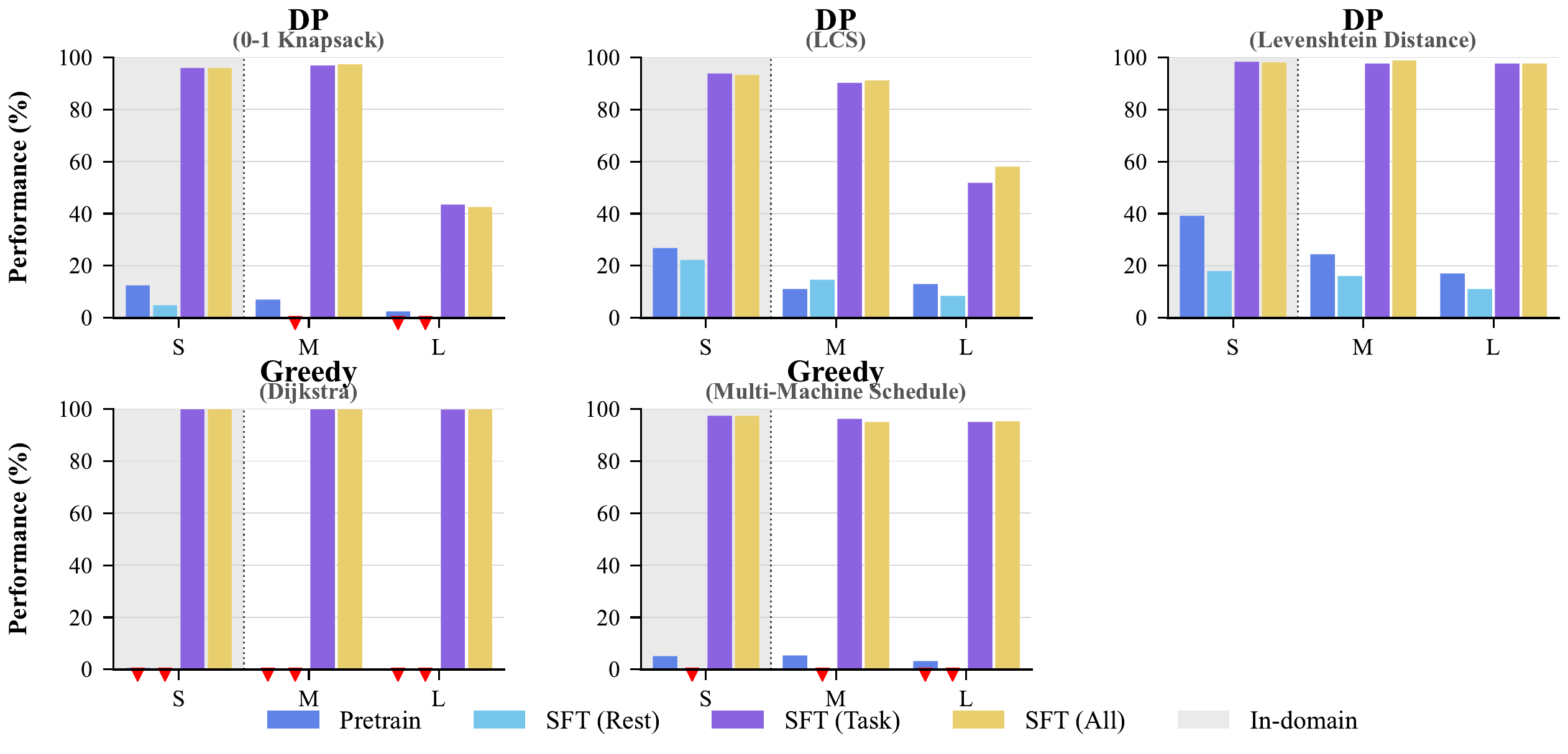}
    \caption{Generalization performance between tasks within a single algorithm (\eg, DP and Greedy). \textbf{Pretrain} represents Qwen 2.5-7B as the basis for subsequent SFT. For each task, \textbf{SFT(Rest)} indicates training using data from other tasks within the algorithm, \textbf{SFT(Task)} indicates training using data from this task, and \textbf{SFT(All)} indicates training using data from all tasks within this algorithm. \textcolor{red}{$\blacktriangledown$} indicates that this piece of data has a label accuracy of less than 5\%.} 
    \vspace{-10pt}
    \label{fig:task_gen}
\end{figure}

For tasks that belong to the same algorithmic idea, we test whether they have combinatorial generalization (\ie, generalization between tasks). As shown in Figure~\ref{fig:task_gen}, the combinatorial generalization property is not significant, which is the target of our future works.

% \clearpage

\setcounter{table}{0} 
\setcounter{figure}{0}
\setcounter{equation}{0}
\renewcommand{\thetable}{J\arabic{table}}
\renewcommand\thefigure{J\arabic{figure}} 
\renewcommand\theequation{J\arabic{equation}}

\section{CoT Length Comparison}
\label{sec:length_cmp}

Since \methodname{} simulates the implementation of a Turing machine, where all algorithm steps are expanded and explicit recall of operands is added, it may lead to a significant increase in CoT. Therefore, taking \textit{Compare Number} task of \textit{Simulation} algorithm as an example, we compared the average lengths of TAIL-CoT and DeepSeek-R1 CoT.

As shown in Table~\ref{tab:tail_vs_deepseek}, despite the increased length, the average number of tokens of \methodname{} remains on par with that of DeepSeek-R1. With such acceptable length, it significantly outperforms the 671B reasoning model with only 7B parameters, which demonstrates the efficiency of \methodname{}.

\begin{table}[h]
\centering
\resizebox{0.72\linewidth}{!}{
\begin{tabular}{l|c|c}
\toprule[1pt]
\multicolumn{1}{c|}{\textbf{Metric}} & \multicolumn{1}{c|}{\textbf{\methodname{}-CoT (7B, Finetuned Qwen2.5 7B)}} & \multicolumn{1}{c}{\textbf{DeepSeek-R1 (671B)}} \\
\midrule[0.5pt]
avg. Tokens & 1455 & 1461 \\
Label Accuracy & 90.0 & 51.2 \\
\bottomrule[1pt]
\end{tabular}
}
\vspace{1em}
\caption{Comparison between \methodname{}-CoT (7B, Finetuned Qwen2.5 7B) and DeepSeek-R1 (671B) on average token usage and label accuracy. The average number of tokens retains the integer part.}
\label{tab:tail_vs_deepseek}
\vspace{-1em}
\end{table}

\setcounter{table}{0} 
\setcounter{figure}{0}
\setcounter{equation}{0}
\renewcommand{\thetable}{K\arabic{table}}
\renewcommand\thefigure{K\arabic{figure}} 
\renewcommand\theequation{K\arabic{equation}}

\section{DeepSeek-R1-Distill Training Results}
\label{sec:distill_cmp}

For a fairer comparison, we fine-tuned Qwen2.5-7B using the same amount of DeepSeek-R1-Distill data and \methodname{}-CoT. As shown in Table~\ref{tab:tail_vs_r1distill}, although DeepSeek-R1's CoT can significantly improve in-domain performance, the improvement is not as significant as \methodname{}-CoT, and the performance shows a sharp decline in longer sequences. Since DeepSeek-R1 has a low accuracy rate on some tasks, distilling the same amount of training data requires a large number of tokens, so this experiment was not conducted on a large scale.

\begin{table}[h]
\centering
\resizebox{0.5\linewidth}{!}{
\begin{tabular}{l|c|c}
\toprule[1pt]
\multicolumn{1}{c|}{\textbf{Setting}} & \multicolumn{1}{c|}{\textbf{TAIL (7B)}} & \multicolumn{1}{c}{\textbf{R1-Distill-Qwen2.5-7B}} \\
\midrule[0.5pt]
S (In Domain) & 98.0 & 72.2 \\
M            & 94.2 & 67.2 \\
L              & 90.0 & 61.8 \\
\bottomrule[1pt]
\end{tabular}
}
\vspace{1em}
\caption{Performance comparison between \methodname{} (7B) and R1-Distill-Qwen2.5-7B across different sequence lengths. (\textbf{S} = Short sequence data, \textbf{M} = Medium sequence data, \textbf{L} = Long sequence data)}
\label{tab:tail_vs_r1distill}
\vspace{-1em}
\end{table}

% \clearpage

\setcounter{table}{0} 
\setcounter{figure}{0}
\setcounter{equation}{0}
\renewcommand{\thetable}{L\arabic{table}}
\renewcommand\thefigure{L\arabic{figure}} 
\renewcommand\theequation{L\arabic{equation}}
\section{CoT Struture Comparison of DeepSeek-R1 and \methodname{}}
\label{sec:cot_cmp}

Take a \textit{Word Flip} task of \textit{Simulation} algorithm as an example. 

\textbf{Zero-shot Query}: Provide the flipped string based on "As he measured, he said, Every Ollivander wand has a core of a powerful magical substance". Reverse the entire string character by character (including spaces).

\lstdefinestyle{llmresponse}{
    language=,
    basicstyle=\ttfamily\small,   % 等宽字体
    backgroundcolor=\color{gray!10},
    frame=single,
    breaklines=true,              % 自动换行
    columns=fullflexible,
    keepspaces=false,
    escapeinside={(*@}{@*)}       % 定义转义岛
}

\subsection{Response of DeepSeek-R1}
\label{res_dsr1_cot}

\begin{lstlisting}[style=llmresponse]
<think>
We are to flip the string based on the given sentence:  "As he measured, he said, Every Ollivander wand has a core of a powerful magical substance"
The instruction is to "flip" the string. Let me reverse the entire string: 
Original: "As he measured, he said, Every Ollivander wand has a core of a powerful magical substance" We can reverse it character by character. 
Steps: 
1. Take the string. 
2. Reverse the order of the characters. 
Example: "hello" becomes "olleh" 
Let's do it: 
Original:    'A','s',' ','h','e',' ','m','e','a','s','u','r','e','d',',',' ','h','e',' ','s','a','i','d',',',' ','E','v','e','r','y',' ','O','l','l','i','v','a','n','d','e','r',' ','w','a','n','d',' ','h','a','s',' ','a',' ','c','o','r','e',' ','o','f',' ','a',' ','p','o','w','e','r','f','u','l',' ','m','a','g','i','c','a','l',' ','s','u','b','s','t','a','n','c','e' 
Reverse:    
Start from the last character and go to the first. 
But note: we can use a simple method in programming to reverse the string. 
However, since we are to do it manually, I'll write the reversed string. 
Let me break the original string into parts and then reverse the whole: 
Original: "As he measured, he said, Every Ollivander wand has a core of a powerful magical substance" 
We can write the reversed string as: 
Last character: 'e' Then: 'c', 'n', 'a', 't', 's', 'b', 'u', 's', ' ', 'l', 'a', 'c', 'i', 'g', 'a', 'm', ' ', 'l', 'u', 'f', 'r', 'e', 'w', 'o', 'p', ' ', 'a', ' ', 'f', 'o', ' ', 'e', 'r', 'o', 'c', ' ', 'a', ' ', 's', 'a', 'h', ' ', 'd', 'n', 'a', 'w', ' ', 'r', 'e', 'd', 'n', 'a', 'v', 'i', 'l', 'l', 'O', ' ', 'y', 'r', 'e', 'v', 'E', ' ', ',', 'd', 'i', 'a', 's', ' ', 'e', 'h', ' ', ',', 'd', 'e', 'r', 'u', 'a', 's', 'e', 'm', ' ', 'e', 'h', ' ', 's', 'A' 
But wait, let's do it step by step: 

...... (*@\textcolor{blue}{(A lot of splitting to achieve the number of digits omitted.)}@*)

Wait, I think I made a mistake. 
Let me write the entire string without spaces and then add the spaces and punctuation: 
Actually, it's easier to just reverse the entire string as a whole. 
We can write the string as: s = "As he measured, he said, Every Ollivander wand has a core of a powerful magical substance" 
Reversed string = s[::-1] in Python. 
But manually: 
Start from the end: 'e' (from substance) 'c' 'n' 'a' 't' 's' 'b' 'u' 's' ' ' (space before substance? 
but actually the last word is "substance", and before that is "magical", so there's a space between them) 

...... (*@\textcolor{blue}{(A lot of repeated verification attempts omitted.)}@*)

Actually, the string is: ... a powerful magical substance So the order in the original is:    ... 'a', space, 'p','o','w','e','r','f','u','l', space, 'm','a','g','i','c','a','l', space, 's','u','b','s','t','a','n','c','e' Therefore, when reversed, we start from the last character of the last word and go backwards: Start: 'e' Then: 'c','n','a','t','s','b','u','s' -> then space -> then the word "magical" reversed: 'l','a','c','i','g','a','m' -> then space -> then the word "powerful" reversed: 'l','u','f','r','e','w','o','p' -> then space -> then 'a' -> then space -> then the word "of" becomes 'f','o'? 

...... (*@\textcolor{blue}{(A lot of attempts at different reversal methods omitted.)}@*)

Actually, the comma we already handled?  Wait, let me reconstruct: The original string:    "As he measured, he said, Every Ollivander wand has a core of a powerful magical substance" Breaking down:   "As" -> ['A','s']   space -> [' ']   "he" -> ['h','e']   space -> [' '] ...

We can reverse the entire list. The last character is 'e' (from substance), then the one before is 'c', then 'n', then 'a', then 't', then ...

Therefore, the flipped string is: "ecnatsbus lacigam lufrewop a fo eroc a sah dnaw rednavillO yrevE ,dias eh ,derusaem e h sA" But note: the original had a space at the end? No, the last word is "substance" without any trailing space. Similarly, the first word "As" has no leading space. So the reversed string should not have a leading space? It ends with "sA", which is correct. Final answer:   "ecnatsbus lacigam lufrewop a fo eroc a sah dnaw rednavillO yrevE ,dias eh ,derusaem e h sA"
</think>

The original string is:  
"As he measured, he said, Every Ollivander wand has a core of a powerful magical substance"

Reversing the entire string character by character (including spaces, punctuation, and maintaining case) results in:  
**"ecnatsbus lacigam lufrewop a fo eroc a sah dnaw rednavillO yrevE ,dias eh ,derusaem e h sA"**

### Explanation:
- The reversal is done on a per-character basis, so the last character of the original string ("e" from "substance") becomes the first character of the flipped string, and so on.
- Spaces, commas, and capitalization are preserved as they appear in the original.
- This is not a word reversal (where words are reversed but characters within words stay in order); it is a full character-level reversal.
\end{lstlisting}

The final output is "ecnatsbus lacigam lufrewop a fo eroc a sah dnaw rednavillO yrevE ,dias eh ,derusaem {\color{red}e h} sA". The model tried many methods, but did not actually perform the flipping operation step by step, and directly output an incorrect result at the end of the inference.

\subsection{\methodname{}-CoT-\textit{styled}}
\label{res_tail}

\begin{lstlisting}[style=llmresponse]
<think>
Decompose the string bit by bit(starting from $1$):
 - Index $1$: 'A'
 - Index $2$: 's'
 - Index $3$: ' '
 - Index $4$: 'h'
 - Index $5$: 'e'
 - Index $6$: ' '
 - Index $7$: 'm'
 - Index $8$: 'e'
 - Index $9$: 'a'
 - Index $10$: 's'
 
 .... (*@\textcolor{blue}{(Omitted and will be expanded in actual reasoning.)}@*)
 
 - Index $80$: ' '
 - Index $81$: 's'
 - Index $82$: 'u'
 - Index $83$: 'b'
 - Index $84$: 's'
 - Index $85$: 't'
 - Index $86$: 'a'
 - Index $87$: 'n'
 - Index $88$: 'c'
 - Index $89$: 'e'

We find that 'As he measured, he said, Every Ollivander wand has a core of a powerful magical substance' has $89$ words, indexed from $1$ to $89$.

The original string is the concatenation of the letters with indices from $1$ to $89$. If you reverse it, the result is the reverse concatenation of the indices, that is, from $89$ to $1$.

Let's perform the reverse stitching step by step:
 - Index $89$: 'e', intermediate result is 'e'
 - Index $88$: 'c', intermediate result is 'ec'
 - Index $87$: 'n', intermediate result is 'ecn'
 - Index $86$: 'a', intermediate result is 'ecna'
 - Index $85$: 't', intermediate result is 'ecnat'
 - Index $84$: 's', intermediate result is 'ecnats'
 - Index $83$: 'b', intermediate result is 'ecnatsb'
 - Index $82$: 'u', intermediate result is 'ecnatsbu'
 - Index $81$: 's', intermediate result is 'ecnatsbus'
 - Index $80$: ' ', intermediate result is 'ecnatsbus '
 
 .... (*@\textcolor{blue}{(Omitted and will be expanded in actual reasoning.)}@*)
 
 - Index $10$: 's', intermediate result is 'ecnatsbus lacigam lufrewop a fo eroc a sah dnaw rednavillO yrevE ,dias eh ,derus'
 - Index $9$: 'a', intermediate result is 'ecnatsbus lacigam lufrewop a fo eroc a sah dnaw rednavillO yrevE ,dias eh ,derusa'
 - Index $8$: 'e', intermediate result is 'ecnatsbus lacigam lufrewop a fo eroc a sah dnaw rednavillO yrevE ,dias eh ,derusae'
 - Index $7$: 'm', intermediate result is 'ecnatsbus lacigam lufrewop a fo eroc a sah dnaw rednavillO yrevE ,dias eh ,derusaem'
 - Index $6$: ' ', intermediate result is 'ecnatsbus lacigam lufrewop a fo eroc a sah dnaw rednavillO yrevE ,dias eh ,derusaem '
 - Index $5$: 'e', intermediate result is 'ecnatsbus lacigam lufrewop a fo eroc a sah dnaw rednavillO yrevE ,dias eh ,derusaem e'
 - Index $4$: 'h', intermediate result is 'ecnatsbus lacigam lufrewop a fo eroc a sah dnaw rednavillO yrevE ,dias eh ,derusaem eh'
 - Index $3$: ' ', intermediate result is 'ecnatsbus lacigam lufrewop a fo eroc a sah dnaw rednavillO yrevE ,dias eh ,derusaem eh '
 - Index $2$: 's', intermediate result is 'ecnatsbus lacigam lufrewop a fo eroc a sah dnaw rednavillO yrevE ,dias eh ,derusaem eh s'
 - Index $1$: 'A', intermediate result is 'ecnatsbus lacigam lufrewop a fo eroc a sah dnaw rednavillO yrevE ,dias eh ,derusaem eh sA'

</think>

The final result is 'ecnatsbus lacigam lufrewop a fo eroc a sah dnaw rednavillO yrevE ,dias eh ,derusaem eh sA'.

\end{lstlisting}

\end{document}